\documentclass[journal,twoside,web]{ieeecolor}
\usepackage{generic}
\usepackage{cite}
\usepackage{amsmath,amssymb,amsfonts}
\usepackage{algorithmic}
\usepackage{graphicx}
\usepackage{textcomp}
\usepackage{multirow}
\def\BibTeX{{\rm B\kern-.05em{\sc i\kern-.025em b}\kern-.08em
    T\kern-.1667em\lower.7ex\hbox{E}\kern-.125emX}}
\begin{document}
\title{Automatic segmentation of meniscus based on MAE self-supervision and point-line weak supervision paradigm}
\author{Yuhan Xie, Kexin Jiang, Zhiyong Zhang, Shaolong Chen, Xiaodong Zhang and Changzhen Qiu
\thanks{Manuscript received May 8, 2022. This work was supported in part by the Science and Technology Planning Project of Guangdong Science and Technology Department under Grant Guangdong Key Laboratory of Advanced IntelliSense Technology (2019B121203006), it was also supported in part by Department of Medical Imaging, The Third Affiliated Hospital of Southern Medical University (Academy of Orthopedics· Guangdong Province), Guangzhou, China. Yuhan Xie and Kexin Jiang contributed equally to this work. (Corresponding authors: Changzhen Qiu and Xiaodong Zhang) }
\thanks{Yuhan Xie, Zhiyong Zhang, Shaolong Chen and Changzhen Qiu are with School of Electronics and Communication Engineering, Sun Yat-sen University, Guangzhou, China(e-mail: qiuchzh@mail.sysu.edu.cn) }
\thanks{Kexin Jiang and Xiaodong Zhang are with Department of Medical Imaging, The Third Affiliated Hospital of Southern Medical University (Academy of Orthopedics· Guangdong Province), Guangzhou, China(e-mail: )}
}

\maketitle

\begin{abstract}
Medical image segmentation based on deep learning is often faced with the problems of insufficient data sets and long time-consuming labeling. In this paper, we introduce the self-supervised method MAE(Masked Autoencoders) into knee joint images to provide a good initial weight for the segmentation model and improve the adaptability of the model to small datasets. Secondly, we propose a weakly supervised paradigm for meniscus segmentation based on the combination of point and line to reduce the time of labeling. Based on the weak label ,we design a region growing algorithm to generate pseudo-label. Finally we train the segmentation network based on pseudo-labels with weight transfer from self-supervision. Sufficient experimental results show that our proposed method combining self-supervision and weak supervision can almost approach the performance of purely fully supervised models while greatly reducing the required labeling time and dataset size.
\end{abstract}


\section{Introduction}
\label{sec:introduction}
\IEEEPARstart{M}eniscus medical image segmentation \cite{seg1,seg2,seg3,seg4,seg5,seg6,seg7} is an important part of human knee joint health analysis, and it has important guiding significance for subsequent diagnostic analysis \cite{clc1,clc2,clc3}. With the development of deep learning, although fully supervised medical image segmentation has been widely studied, due to the fact that datasets in the medical field are often difficult to obtain or the scale is small, and obtaining high-quality segmentation labels often requires time-consuming production by experienced physicians, it is difficult to train excellent medical image segmentation models with full supervision. In the field of natural images, pre-training transfer learning has been widely studied to improve the generalization ability of the model in downstream tasks in small datasets. However, many related works often rely on a large amount of raw data, such as the Imagenet \cite{imagenet} dataset in the natural field. This is also unacceptable for the medical field. The recently proposed MAE self-supervised paradigm, with its data-enhanced feature of random masking itself, makes it possible to perform self-supervised learning only on a small amount of raw data. However, this paradigm has not received sufficient attention in the medical field, especially in the knee joint field.

In the field of natural images, weakly supervised semantic segmentation has also been fully studied to reduce the time required for labeling, because natural images have very clear boundary characteristics and color discrimination, so weak labels can be used to generate very high-quality pseudo-labels , and used for segmentation tasks. However, in the field of medical images, especially MRI images, there are difficulties such as lack of color attributes, blurred or missing boundaries, and interference of lesion areas. Even robust algorithms such as level sets \cite{li2010distance} often do not work well in MRI images. Therefore, how to design an efficient pseudo-label generation algorithm is the key to applying weak supervision to medical images.

Based on the above problems, we designed a segmentation algorithm based on self-supervision and weak supervision on a meniscus datasets of small scale. The MAE self-supervision paradigm is adopted to provide the model good initial representation ability and generate pretrained weights for the weak supervision; the weak supervision paradigm based on points and lines is adopted to reduce the time and difficulty of labeling. We verify the feasibility and robustness of the algorithm through experiments. Specifically, our contributions are as follows:

1. We apply the MAE self-supervised paradigm to the knee joint dataset, so that the network can also have excellent visual representation capabilities in advance; at the same time, we explore different types of transfer positions based on encoders and deep semantic interaction modules.

2. Based on the structural prior information of the meniscus, we designed a weakly supervised label based on the combination of points and lines, and proposed a backbone-based region growing algorithm to complete the generation from weak labels to pseudo labels. Finally, we explore the most suitable segmentation model in the meniscus dataset based on self-supervised weight transfer and pseudo-labels, and compare with some weakly supervised segmentation networks based on rectangular boxes. Experiments verify our proposed algorithms can almost approximate fully supervised model performance.

\section{Related works}

\subsection{Self supervision}
Self-supervision is an important research direction in current computer vision. It adopts self-supervised tasks that do not rely on additional labels, which enables the model to have good representational capabilities and leads to improvement for related downstream tasks. Self-supervision in the visual field is mostly divided into two types of tasks: contrastive learning and reconstruction. Among them, contrastive learning aims to minimize the gap of similar features and maximize the gap of heterogeneous features. The classic work includes SimCLR \cite{simclr}, etc.; while the reconstruction tasks are mostly based on the original image. The derived image is the input, and the original image or its features are predicted, and related methods include MAE \cite{mae}, Simmim \cite{simmim}, Maskfeat \cite{maskfeat} and so on. Among them, MAE has received sufficient attention for its simple and efficient architecture. It takes randomly masked images as input and reconstructs the masked image regions, thereby exploiting the potential of the model to utilize contextual reasoning and greatly improving the visual representation ability of the model.

\subsection{Weak supervision}
Weak supervision is a method that utilizes weak labels whose semantic information is weaker than standard labels to complete related tasks. In the visual domain, the classical weak label forms for semantic segmentation are: rectangular boxes \cite{box1,box2,box3,box4}, point and scribbles \cite{point,scribble1,scribble2,scribble3,scribble4}, and image-level labels \cite{image1,image2,image3,image4,image5}. Most of the algorithms based on rectangular boxes use traditional algorithms such as GrabCut\cite{gradcut}, DenseCRF \cite{densecrf}, etc. to generate pseudo-labels, and train segmentation model based on pseudo-labels and rectangular boxes. Representative works include BCM \cite{box1}, Box2Seg \cite{box4} etc.; the algorithm based on points or scribbles outlines the main position/shape of the target area with points and scribbles, however, due to the lack of strict boundary information like a rectangular box, the performance is generally worse than that of the rectangular box-based method; image-level label based methods relies solely on image-level annotations, such as object categories possessed in the image, for pseudo-label generation. Classical methods are mostly based on encoders to generate class activation maps (CAMs) \cite{cam}, which then generate regions of interest for segmentation. This method requires the simplest labeling process, but because its weak labels provide the least amount of information, its performance is poor compared to other methods.

\subsection{Full supervision}
Full supervision is to use complete real labels to complete the training. In medical image segmentation, fully supervised segmentation models have been well explored. In terms of network module components, it can be roughly divided into CNN-based models and transformer-based models. In the design of medical image segmentation models, most of them follow the design architecture of standard encoders and decoders, such as the most classic unet \cite{unet}. In addition, unet++ \cite{unet++}, unet3+ \cite{unet3+} explored the skip connection method of the network; while resunet \cite{resunet} carried out a mixed design of convolution and residual structure to improve the performance of the convolution module; attunet \cite{attunet} introduced a gated attention mechanism based on unet, which facilitated more effective fusion of encoder and decoder features. Among the models based on transformer \cite{transformer}, transunet \cite{transunet} introduced the module of VIT \cite{vit} into the unet architecture for the first time for deep semantic interaction, while swinunet \cite{swinunet} replaced the convolution module with a pure transformer form. In addition, efficient 3D segmentation networks such as UNETR \cite{unetr} and nnformer \cite{nnformer} have also been proposed one after another, which also showed good performance in 3D image segmentation.

\section{Methods}

\subsection{Overall structure}
Our overall framework for meniscus segmentation is mainly composed of self-supervised pre-training and weakly-supervised segmentation network training. The overall structure is shown in the figure 1:

\begin{figure}[!t]
	\centerline{\includegraphics[width=\columnwidth]{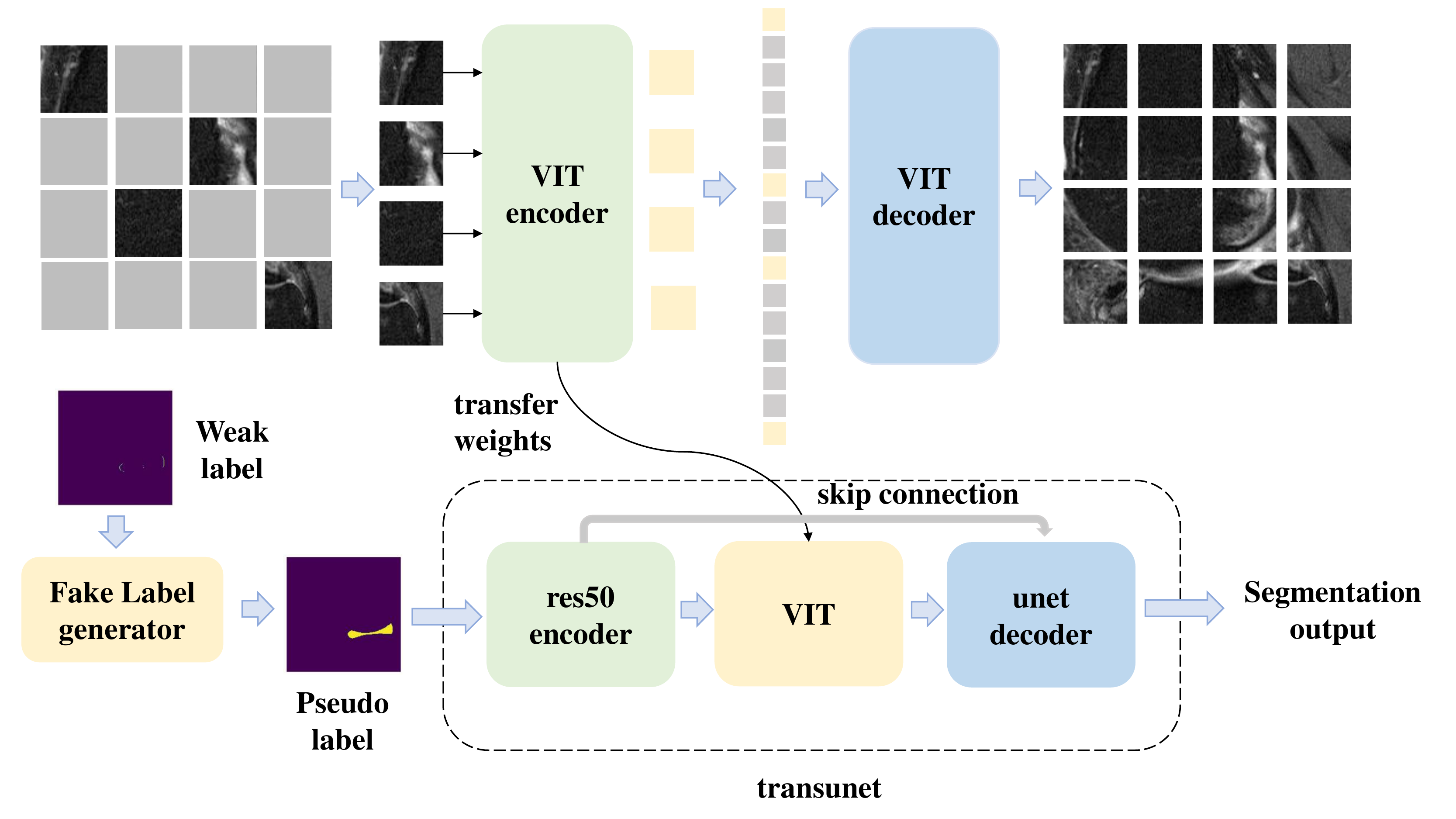}}
	\caption{Self-Supervision + Weak Supervision Overall Framework}
	\label{fig1}
\end{figure}

The self-supervised part adopts the structure of encoder and decoder, and the pre-training of the encoder is only completed based on the original MRI image data; in the weakly supervised part, we first use a point-line-based region growing algorithm to generate pseudo-labels from weak labels. The interaction layer of the segmentation network is initialized with pretrained encoder weights, and then the segmentation network is trained according to the pseudo-labels. The overall framework can achieve almost fully supervised performance based only on the original image data and weak labels of points and lines.

\subsection{Self supervision}\label{formats}
In the self-supervised task, we adopt the training architecture of MAE to predict the original image from the randomly occluded images, resulting in a good visual representation of the meniscus dataset and a good initialization weight for the next weak supervision. The overall structure is mainly divided into the following three parts:

\textbf{Preprocessing}: In the MAE architecture, the preprocessing part mainly completes the conversion from two-dimensional images to randomly masked sequences, including patch embedding , position embedding and random masking. Its formula expression is as follows:

\begin{equation}
	X_{p a}=P a E m b(X) ; X \in R^{H \times W}, X_{p a} \in R^{\frac{H W}{P^{2} \times D}}
\end{equation}
\begin{equation}
	\begin{split}
	&X_{p o s}=\operatorname{concat}\left(X_{p a}, X_{c l c}\right)+E_{p o s} ;\\
	& X_{c l c} \in R^{1 \times D}, E_{p o s} \in R^{\left(\frac{H W}{P^{2}+1}\right) \times D}
	\end{split}
\end{equation}
\begin{equation}
	X_{\text {mask }}=\operatorname{mask}\left(X_{\text {pos }}\right) ; X_{\text {mask }} \in R^{\operatorname{ratio}\left(\frac{H W}{P^{2}+1}\right) \times D}
\end{equation}

A P*P image block is set as the basic unit in the MAE task. We describe the input two-dimensional image slice size as X, after patch embedding, the serialized feature representation $X_{p a}$ is obtained, where D represents the serialized code length. P=16, D=768, H=W=224, which are consistent with the parameter settings of the VIT-base model in the VIT paper. Then, the clc token and position embedding are added to the output of the patch embedding. Finally, 75 \% of the image blocks are randomly masked in the sequence according to the preset ratio ratio=0.25 of MAE.

\textbf{Encoder}: The encoder uses 12 consecutive basic transformer modules, in which the encoder only encodes the visible 25 \% of the image blocks, so it can greatly save the storage and time overhead in the encoding process, and avoid additional error interference. The expression of the basic module is as follows:

\begin{equation}
	Z_{l}^{\prime}=M S A\left(L N\left(Z_{l-1}^{\prime}\right)\right)+Z_{l-1}
\end{equation}
\begin{equation}
	Z_{l}=M L P\left(L N\left(Z_{l}^{\prime}\right)\right)+Z_{l}^{\prime}
\end{equation}

\textbf{Decoder}: The decoder needs to decode all image blocks, so it is necessary to input the encoder output together with the mask image block into the decoder, and positionally encode all image blocks. After position encoding, the decoder is also composed of 8 consecutive transformer modules stacked, and its output is sent to the loss function for comparison with the original image, in which we only calculate the loss for the masked image blocks. According to the conclusion of MAE, We use the normalized result of the original masked blocks as the prediction target.

\subsection{Weak supervision}

\subsubsection{Pseudo-label generation}
The meniscus can be divided into anterior horn, posterior horn, and body according to its positional relationship. In the sagittal view, the anterior and posterior horns are mostly triangular, while the body is mostly symmetrical double-triangular. Based on the prior information of the meniscus, we summarize the characteristics of the meniscus: there are many fibrous tissues on both sides of the meniscus, so it is difficult to define the boundary of the meniscus; while the boundary on the inner side of the meniscus is mostly clear. Therefore, we believe that conventional weak labels such as bounding boxes can not overcome the difficulty of bilateral ambiguity of the meniscus, methods such as level set and Grabcut cannot generate good pseudo-labels from weak labels. 

\begin{figure}[!t]
	\centerline{\includegraphics[width=\columnwidth]{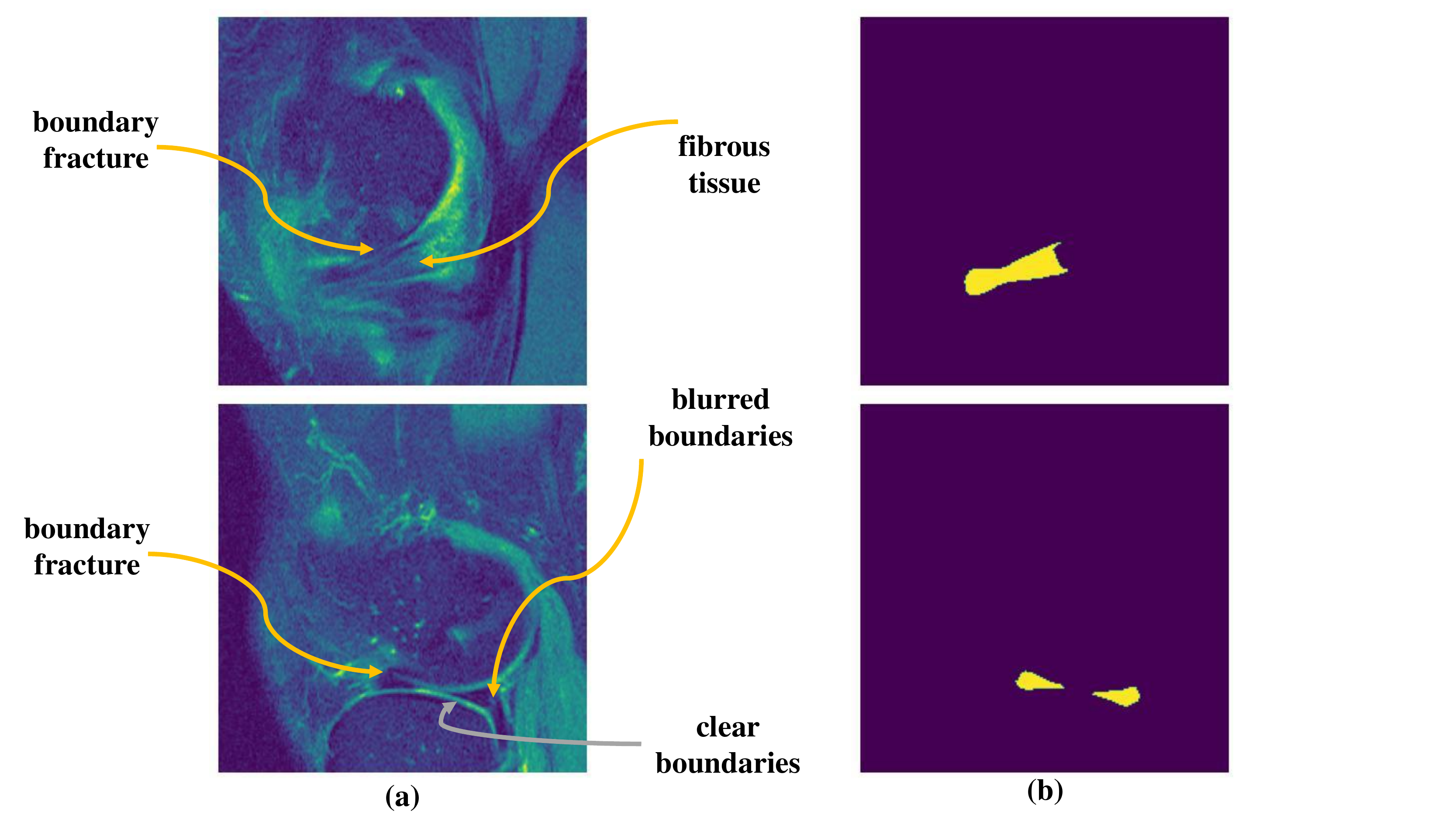}}
	\caption{difficulty in pseudo-label generation (a)original MRI images (b) real label}
	\label{fig2}
\end{figure}

In order to take advantage of the inherent shape and texture prior information of the meniscus, we set the weak labels of point and line with semantic information as follows: The anterior and posterior horns are both annotated with weak labels of one point and one line, while the point represents the corner, the line represents the outer edge of the corner. The body adopts a weak label combination of two points and two lines. The two points respectively represent the upper and lower boundary points at the center of the body, and the two lines respectively represent the edge lines on both sides of the body. 

\begin{figure}[!t]
	\centerline{\includegraphics[width=\columnwidth]{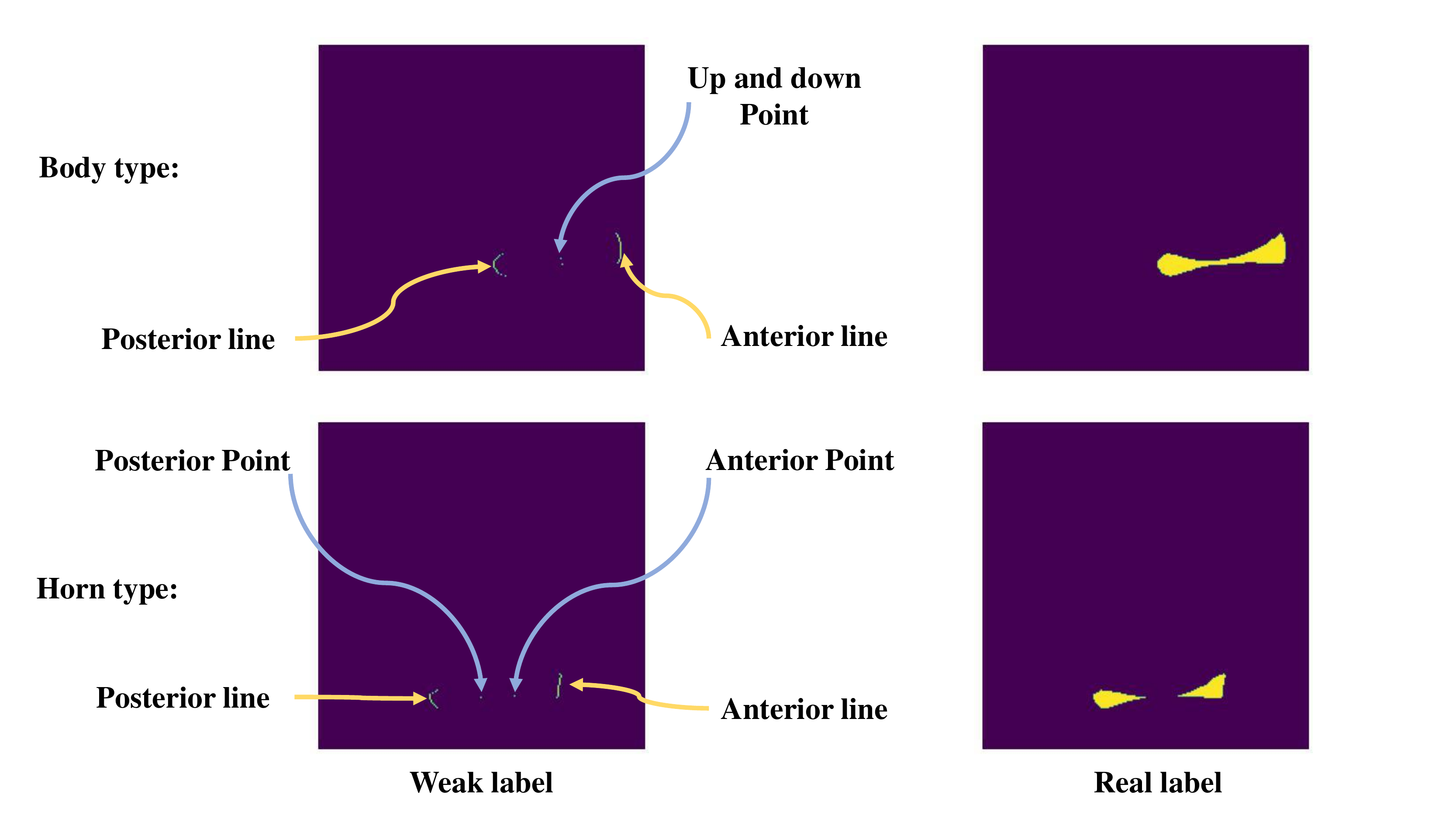}}
	\caption{weak label}
	\label{fig3}
\end{figure}

This weak label form fundamentally leaves the most difficult point of labeling—the determination of the bilateral boundary to the experienced labelers, while the filling of the inner side is automatically generated by the algorithm. In this way, the labeling of points and lines introduces sufficient prior semantic information - the determination of edges into traditional weak labels. At the cost of a small increase in annotation time, the information richness that is difficult to achieve by traditional weak annotation forms is obtained. The algorithm flow of pseudo label generation is as follows:

\textbf{Backbone generation}: We denote the points provided in the left and right horn as $P_{\mathrm{h}}^{p}, P_{\mathrm{h}}^{a}$, and the provided lines as $L_{\mathrm{h}}^{p}, L_{\mathrm{h}}^{a}$; we denote the two points provided in the body as $P_{b}^{u}, P_{b}^{d}$, the provided lines is denoted as$L_{b}^{\mathrm{p}}, L_{b}^{\mathrm{a}}$. The formula for generating the center point is as follows:

\begin{equation}
	P_{\mathrm{h}, c}^{\text {type }}=\frac{L_{\mathrm{h}}^{\text {type }}[\text { mid }]+P_{\mathrm{h}}^{\text {type }}}{2}, \text { type }=\mathrm{a}, \mathrm{p}
\end{equation}
\begin{equation}
	P_{b, c}^{\text {type }}=\frac{L_{b}^{\text {type }}[\text { mid }]+\frac{P_{b}^{u}+P_{b}^{d}}{2}}{2}, \text { type }=\mathrm{a}, \mathrm{p}
\end{equation}

where $L_{\mathrm{h}}^{t y p e}[$ mid $], \quad L_{\mathrm{h}}^{t y p e}[$ mid $]$ represent the midpoint of the horn and body line annotation, $P_{\mathrm{h}, c}^{\text {type }}, P_{b, c}^{\text {type }}$ represents the center point of the horn and body, respectively, and type represents the anterior or posterior side. After getting the center point, we connect the first and last points marked on the line with the center point. If there are double center points(in body type), connect the center point with the center point to obtain the backbone of the body.

\begin{figure}[h]
	\centering
	\begin{minipage}{0.30\linewidth}
		\vspace{3pt}
		\centerline{\includegraphics[width=\textwidth]{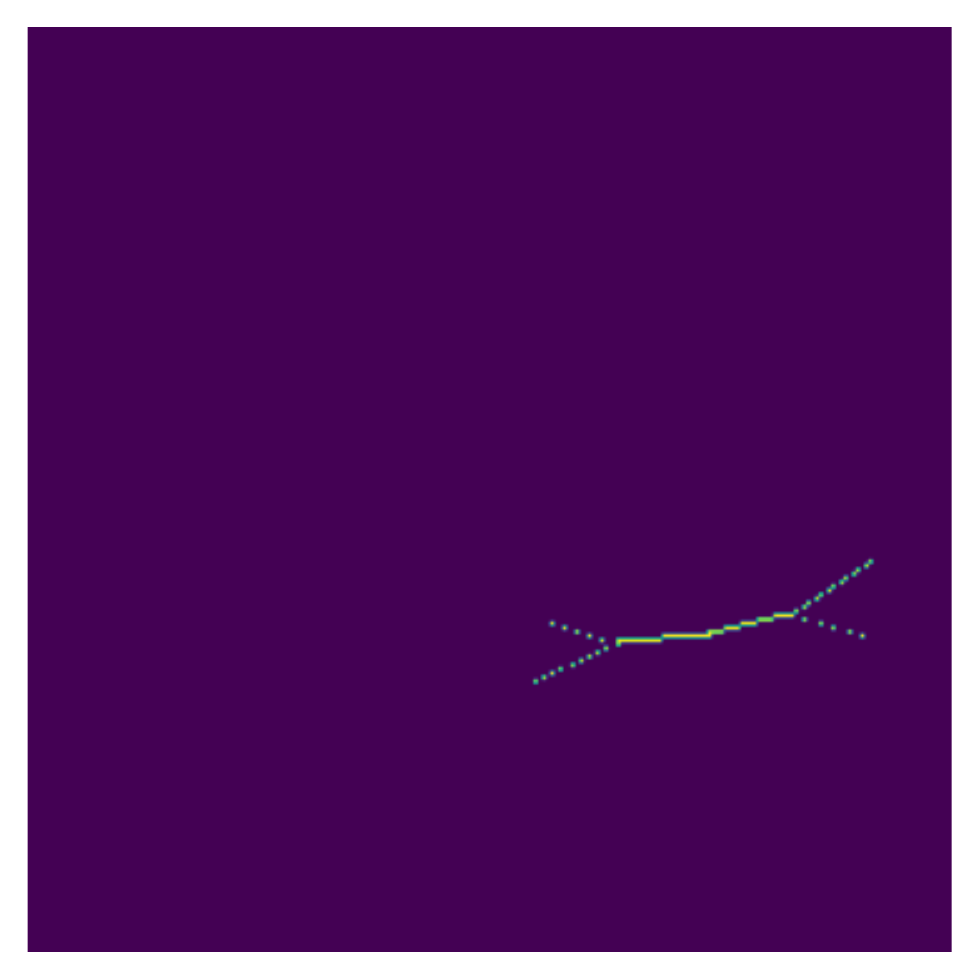}}
	\end{minipage}
	\begin{minipage}{0.30\linewidth}
		\vspace{3pt}
		\centerline{\includegraphics[width=\textwidth]{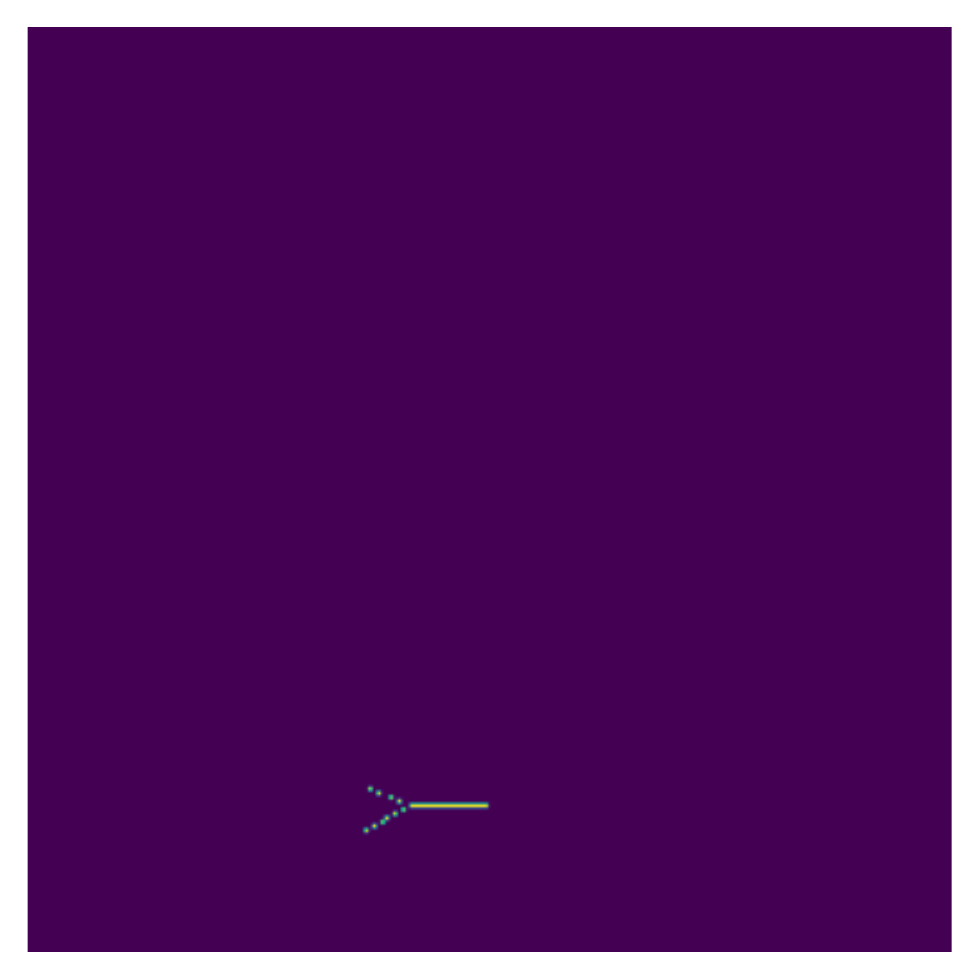}}
	\end{minipage}
	\begin{minipage}{0.30\linewidth}
		\vspace{3pt}
		\centerline{\includegraphics[width=\textwidth]{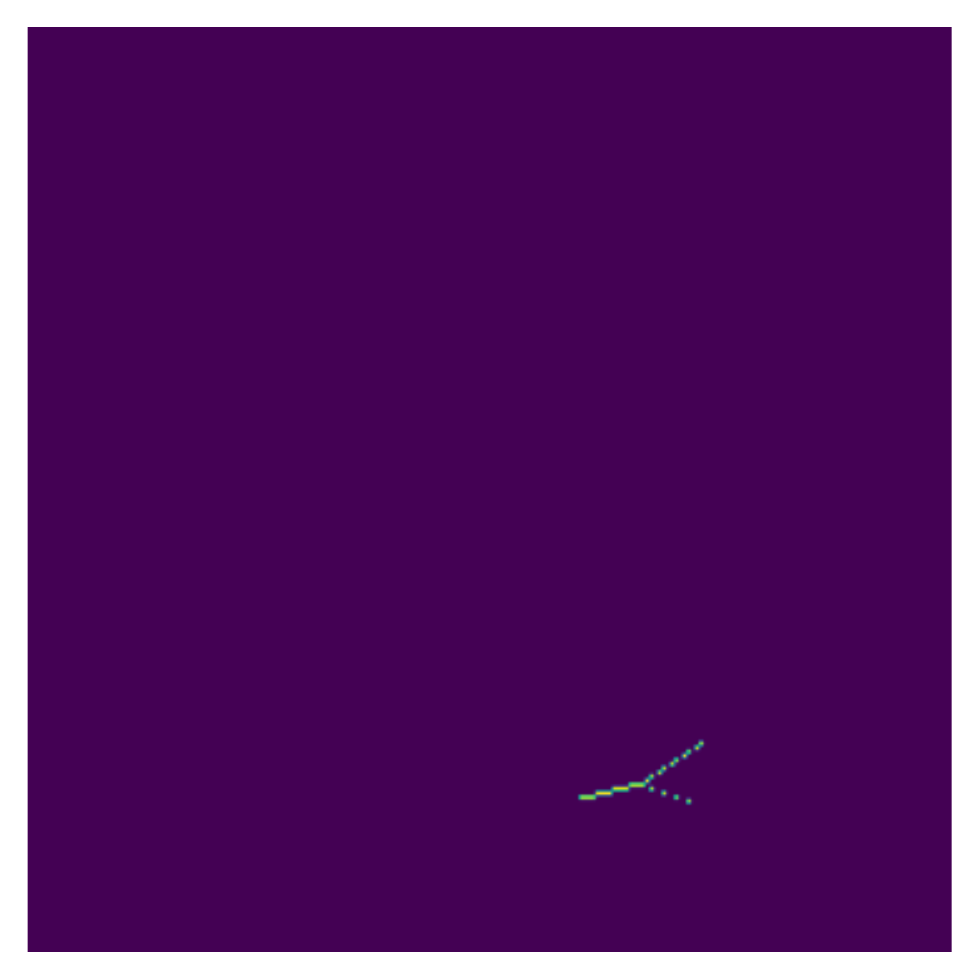}}
	\end{minipage}
	\caption{Visualization of backbone generation; (a) Body backbone (b) Posterior horn backbone (c) Anterior horn backbone}
\end{figure}

\textbf{Difficult area filling}: We connect the head and tail of the line label to the center point, and fill the area enclosed by the double line and the line label to complete the pre-determination of the difficult area, so as to avoid the difficult area in the subsequent area growth. The filling result is shown in the figure 5:

\begin{figure}[h]
	\centering
	\begin{minipage}{0.30\linewidth}
		\vspace{3pt}
		\centerline{\includegraphics[width=\textwidth]{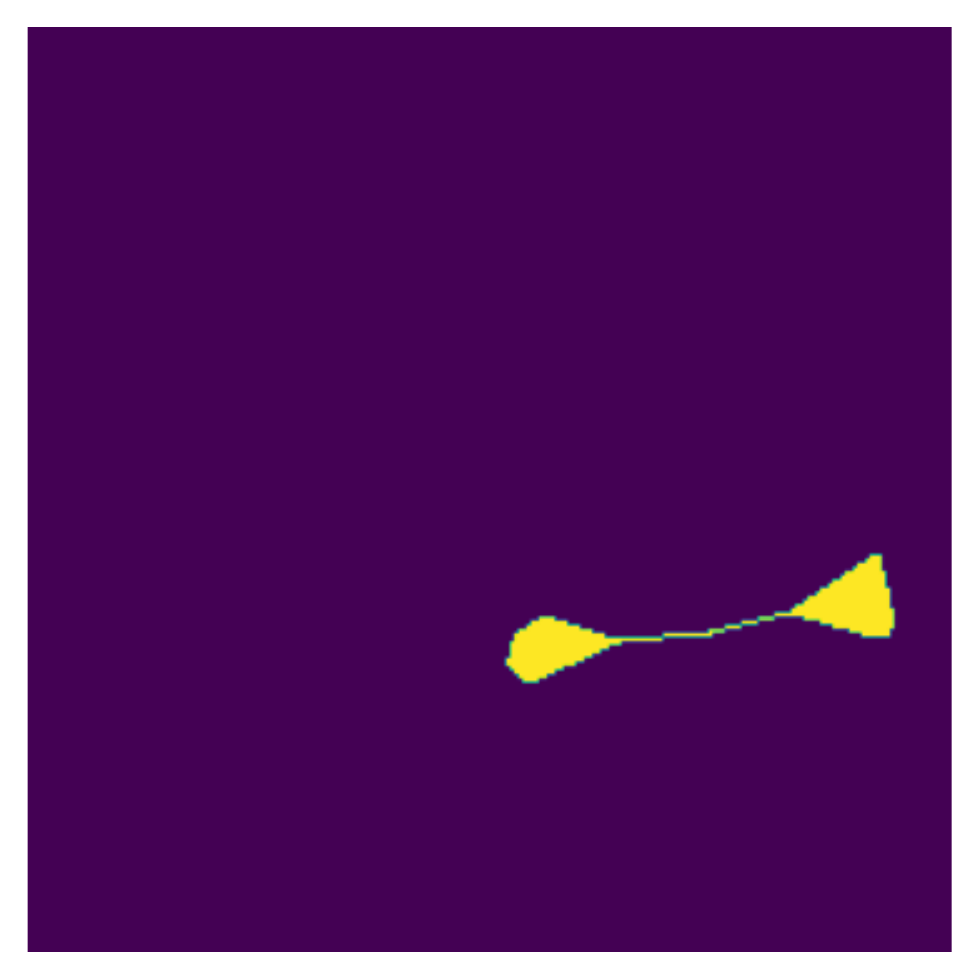}}
	\end{minipage}
	\begin{minipage}{0.30\linewidth}
		\vspace{3pt}
		\centerline{\includegraphics[width=\textwidth]{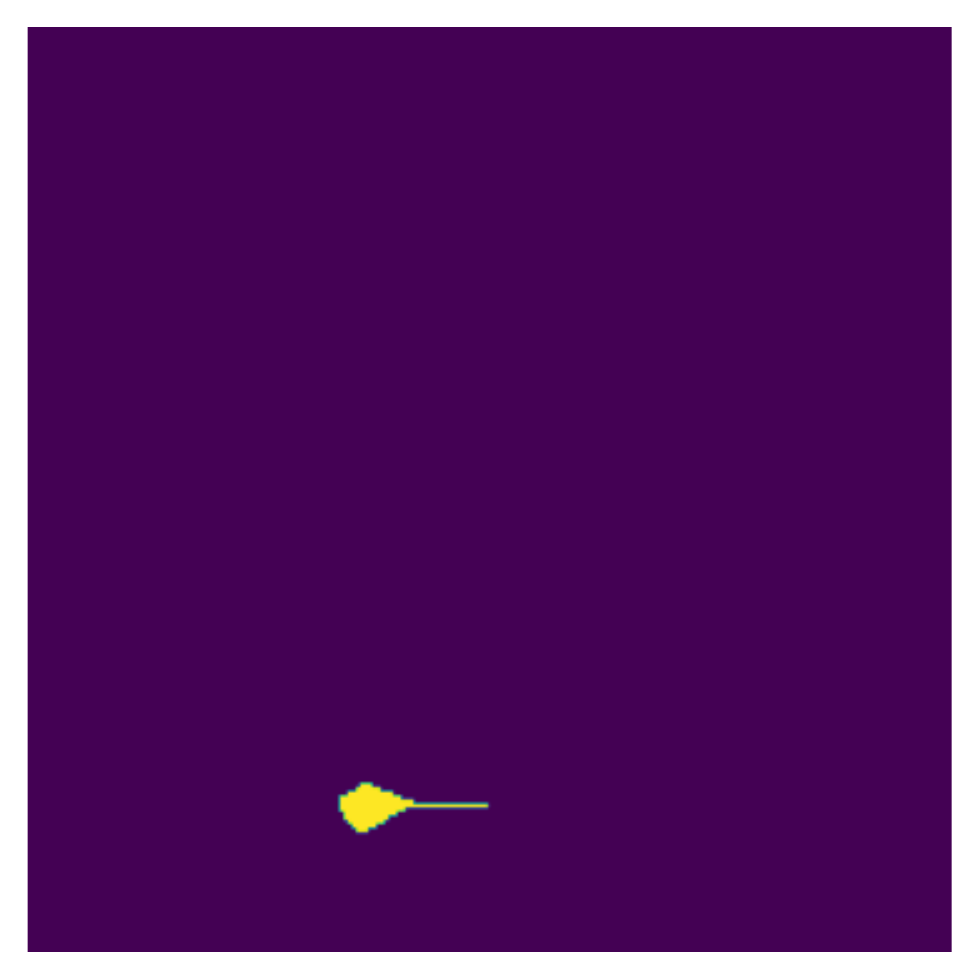}}
	\end{minipage}
	\begin{minipage}{0.30\linewidth}
		\vspace{3pt}
		\centerline{\includegraphics[width=\textwidth]{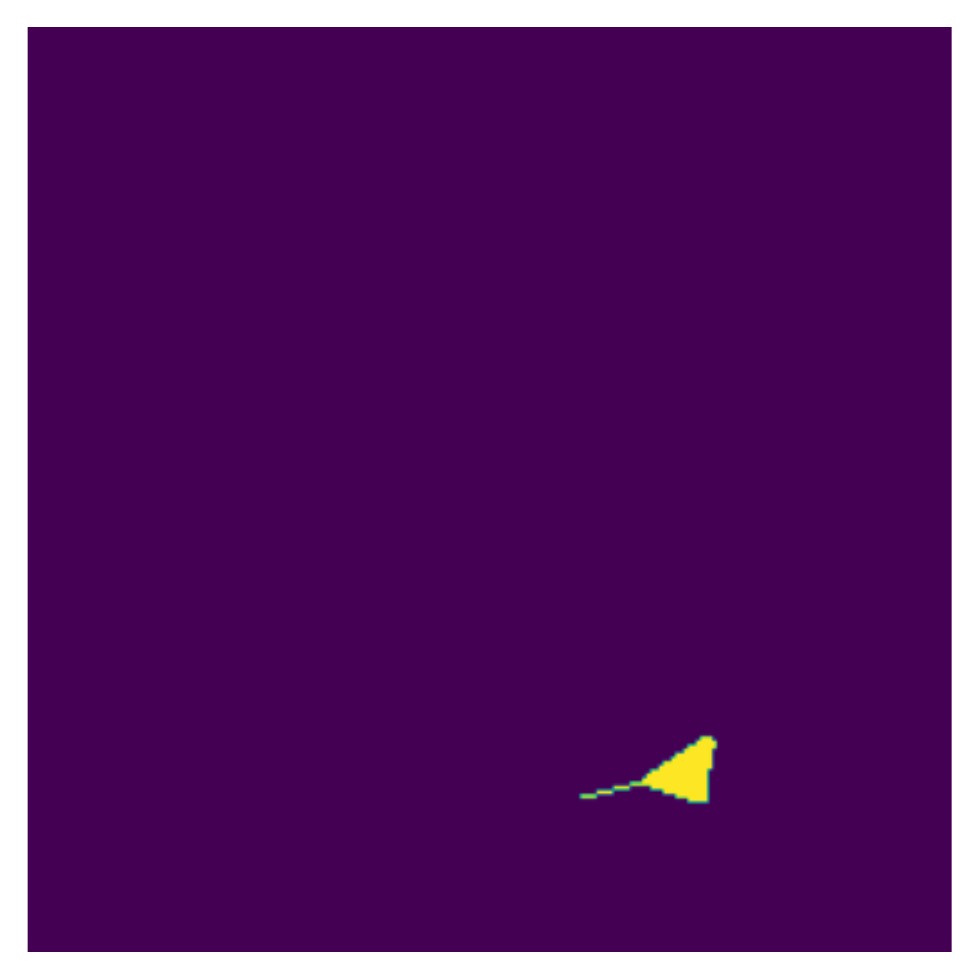}}
	\end{minipage}
	\caption{Visualization of backbone generation; (a) The filling result of the difficult area of the body (b) The filling result of the difficult area of the posterior horn (c) The filling result of the difficult area of the anterior horn}
\end{figure}

\textbf{Regional growth}: We use the points in the backbone as the seed points for regional growth, the filled difficult area as the area that has been grown, and the constraint area formed by the head and tail of the line annotation and the point annotation to constrain. According to the prior information on the shape of the meniscus, we conclude that the following characteristics are: the body is generally wide on both sides and the middle is short, while the horns are wide on the outside and the inside converges into a point. Based on this, we set the maximum range of the boundary of the growth area as shown in figure 6.

\begin{figure}[!t]
	\centerline{\includegraphics[width=\columnwidth]{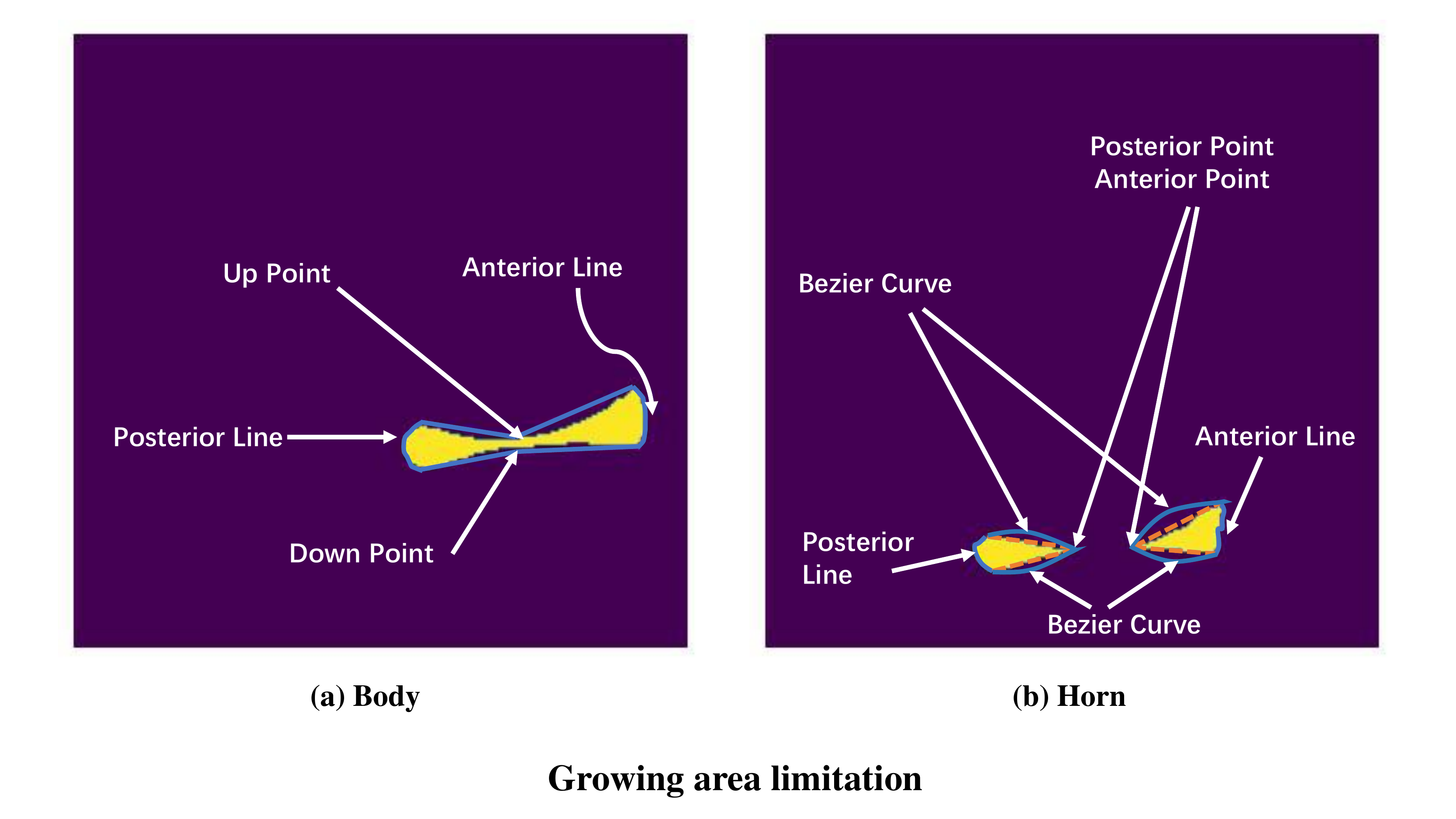}}
	\caption{limitation range of regional growth}
	\label{fig6}
\end{figure}

The inner side of the body is mostly concave, and the boundary can be directly determined by the connection line, while there are some cases in the horn that are convex, so the Bezier curve derived from the connection line is used as the boundary. The control point of the Bezier curve is on the vertical bisector of the midpoint of the connection line, and the vertical distance is set to 3 according to statistical experience, and the corresponding quadratic Bezier curve can be generated according to the first and last points of the connection line and the control point. By correcting the edges, the growth out-of-bounds problem caused by discontinuous edges in MRI images can be sufficiently prevented. The formula for generating a quadratic Bezier curve is as follows:

\begin{equation}
	B(t)=\left(1-t^{2}\right) P_{0}+2 t(1-t) P_{1}+t^{2} P_{2}, t \in[0,1]
\end{equation}

Among them, $P_{0}, P_{2}$ represent the first and last points of the line respectively, and $P_{1}$ represents the control point on the vertical line. Before segmentation, in order to avoid noise points in the meniscus and possible voids caused by tearing high signal, we use mean smoothing to preprocess the image, and then perform region growth. The allowable conditions for zone growth are set as:

\begin{equation}
	\left|P_{g, n e w}-P_{g, a v g_{-} b}\right| \leq \varepsilon
\end{equation}

where $P_{g, n e w}$represents the gray level of the new point, $P_{g, a v g_{-} b}$ represents the average gray level of the backbone point; $\mathcal{E}$ is set to 30 according to statistical experience. After the region grows, continuous inflation and erosion are used to avoid the existence of voids, where the core size of the expansion and corrosion should be consistent to offset the effect on the boundary.

\subsubsection{Pseudo-label generation}
After obtaining the pseudo-labels generated by weak labels, we mainly adopt two kinds of networks for weakly supervised training. Network 1 mainly uses VIT as the encoder and the 2D UNETR decoder as the decoder. The purpose is to best fit the encoder weights provided by MAE self-supervised training, and to examine the performance of the pure transformer encoder in meniscus segmentation. Network 2 adopts the network architecture of Transunet, which is based on resnet50 encoder and unet decoder, and adds a continuous 12-layer transformer module in the middle for global information interaction. The purpose is to investigate the feasibility of the combination of convolution module and transformer module in meniscus segmentation. Therefore, the self-supervised encoder weights are loaded in the position of the standard encoder in network 1, and loaded in the position of the semantic information interaction module (ViT) in network 2. The network structure is shown in figure 7 :

\begin{figure}[!t]
	\centerline{\includegraphics[width=\columnwidth]{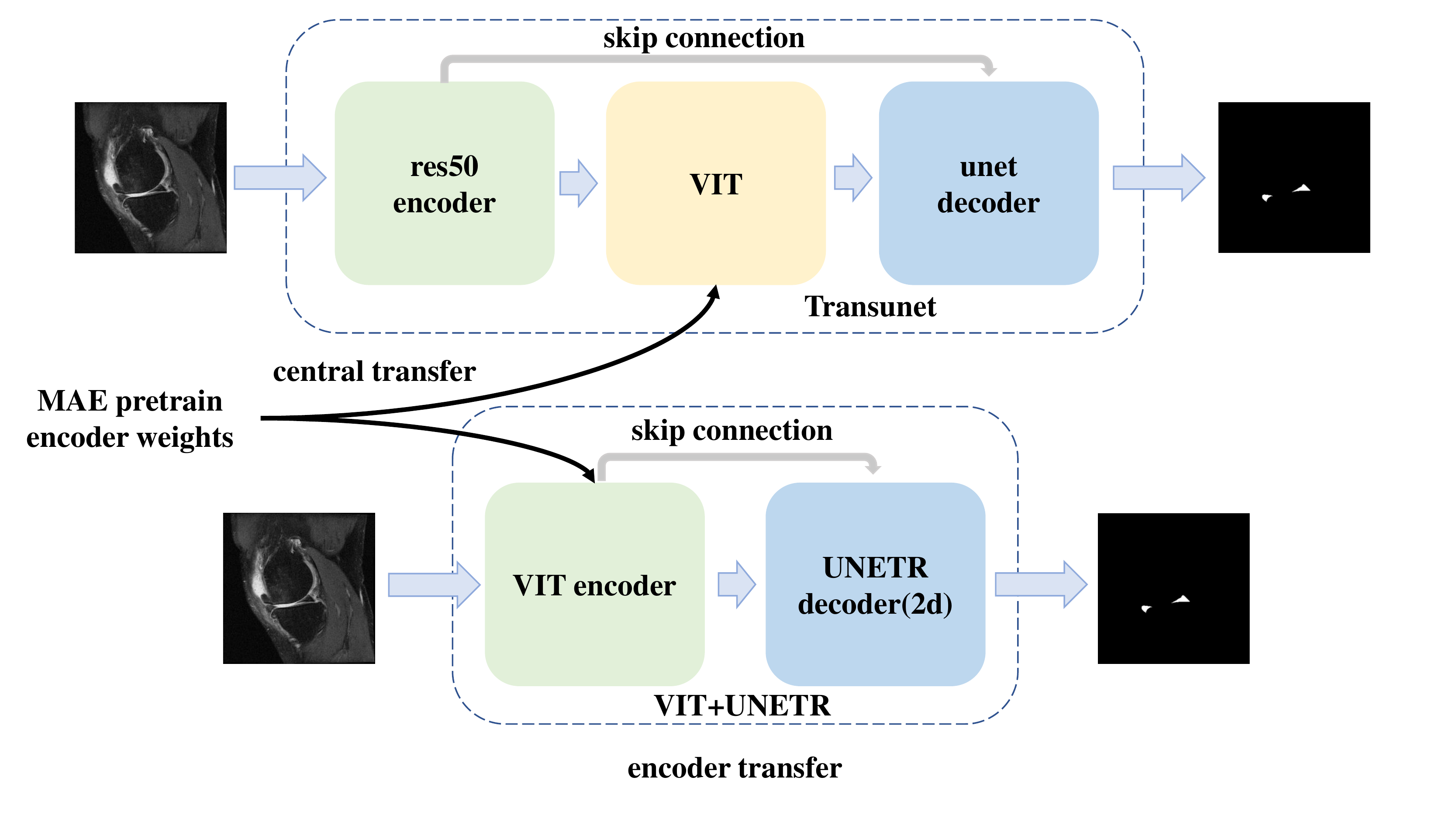}}
	\caption{Schematic diagram of weakly supervised segmentation network}
	\label{fig7}
\end{figure}

The loss function is jointly supervised by the combination of BCE and Dice coefficients, and its expression is as follows:

\begin{equation}
	\operatorname{loss}=\frac{1}{N} \sum_{b=1}^{N}\left(-\frac{1}{2} y_{b} \log \hat{y}_{b}+1-\frac{2 \cdot y_{b} \cdot \hat{y}_{b}}{y_{b}+\hat{y}_{b}}\right)
\end{equation}

where N represents batchsize, $y_{b}$ represents groundtruth, and $\hat{y}_{b}$ represents the prediction result. It is worth mentioning that we noticed that our weak label is also enough to generate a rectangular box containing the label area, so we considered some rectangular box-based SOTA weakly supervised networks in the natural image domain, including Box2Seg, BCM two methods , and the comparison of the specific experimental results will be shown in the follow-up.

\section{Experiments}

\subsection{Dataset, experimental details, evaluation metrics}
In the experiments, we select the partial knee joint dataset of sequence IW-TSE to produce segmentation labels, and conduct self-supervised and weakly supervised experiments based on them. Overall, the dataset is shown in the table 1:

\begin{table}[]
\caption{Basic parameters of the meniscus dataset}
\label{table:headings}
	\begin{center}
		\begin{tabular}{ll}
			\hline
			Parameter           & Index      \\ \hline
			Number              & 110        \\
			Dimensions          & 448*444*37 \\
			Number of 2D slices & 4070       \\ \hline
		\end{tabular}
	\end{center}
\end{table}

Our experiments were completed under the basic environment of RTX 3070, PyTorch1.7, and Python3.7. The basic structure used in self-supervised training is the VIT-Base model, and the relevant parameters are consistent with the original paper of MAE. In the segmentation task, the training set and test set are divided according to the ratio of 7:4, and we crop the image to 224*224 according to statistical experience. Transunet and VIT+UNETR are initialized with the weights generated by self-supervision, and the rest of the models are initialized randomly; during the training process, the batchsize is set to 8, the optimizer is Adam, the learning rate is set to 1e-4, and the weight decay is set to 1e-4 . We use the Dice coefficient to measure the performance of the segmentation model, and its formula is as follows:

\begin{equation}
	D S C=\frac{2(X \cap Y)}{X \cup Y}
\end{equation}

Where X denotes the segmentation output and Y denotes the label.

\subsection{Visualization of MAE reconstruction effect}
\begin{figure}[h]
	\centering
	\begin{minipage}{0.30\linewidth}
		\vspace{3pt}
		\centerline{\includegraphics[width=\textwidth]{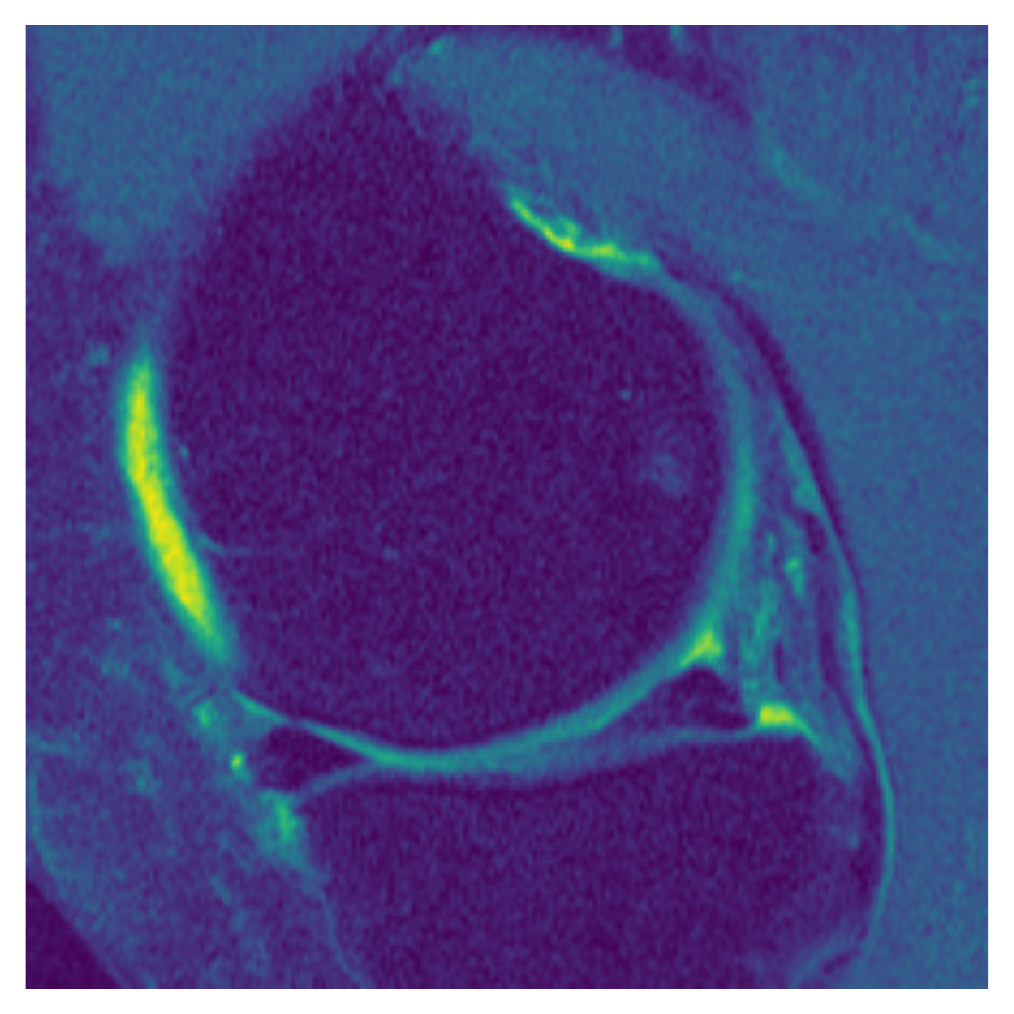}}
		\vspace{3pt}
		\centerline{\includegraphics[width=\textwidth]{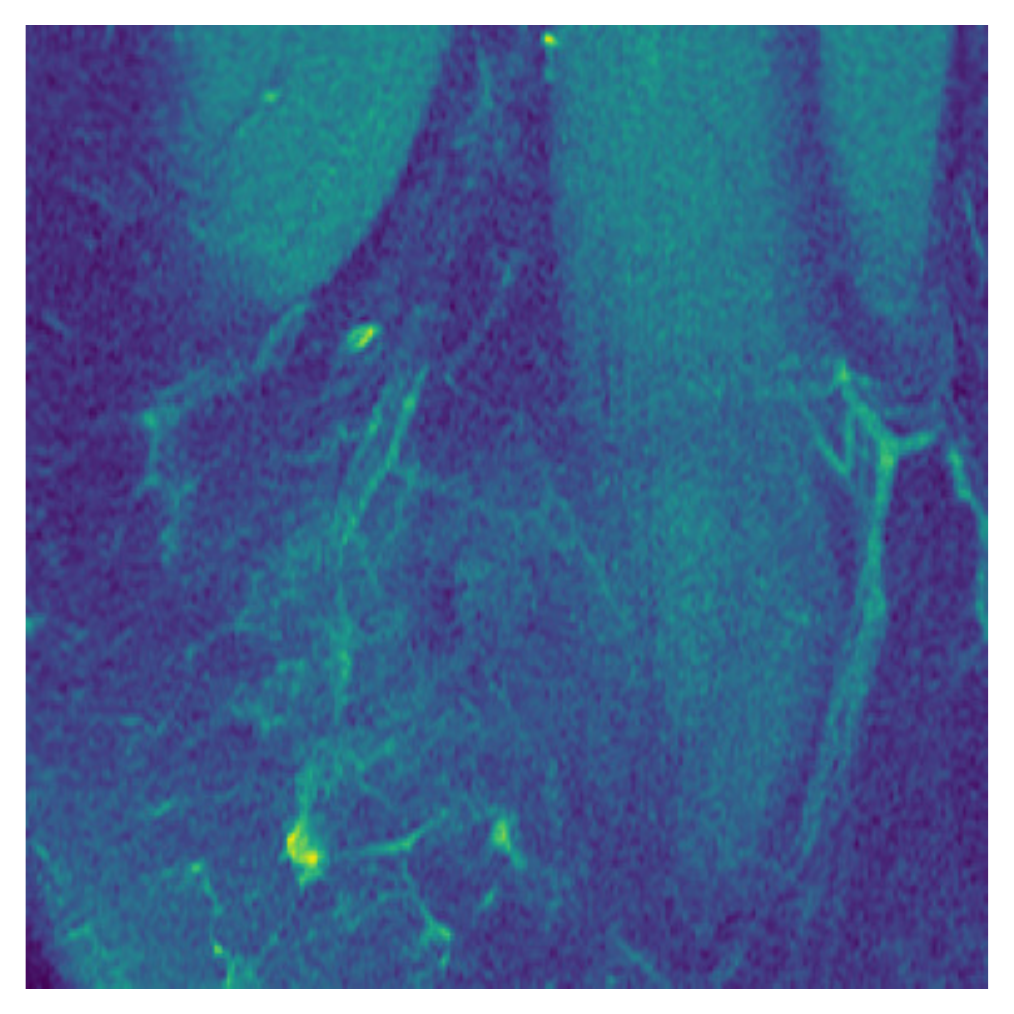}}
		\vspace{3pt}
		\centerline{\includegraphics[width=\textwidth]{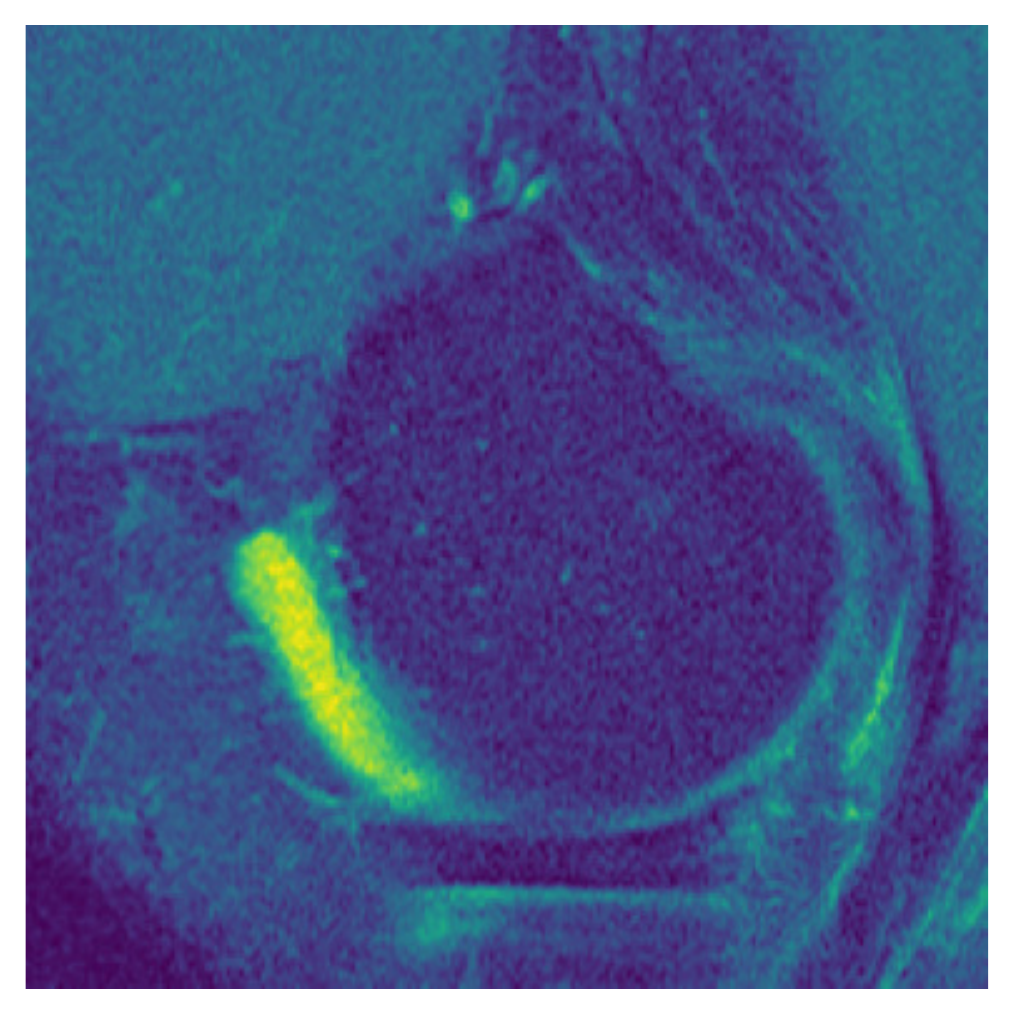}}
		\centerline{(a)}
	\end{minipage}
	\begin{minipage}{0.30\linewidth}
		\vspace{3pt}
		\centerline{\includegraphics[width=\textwidth]{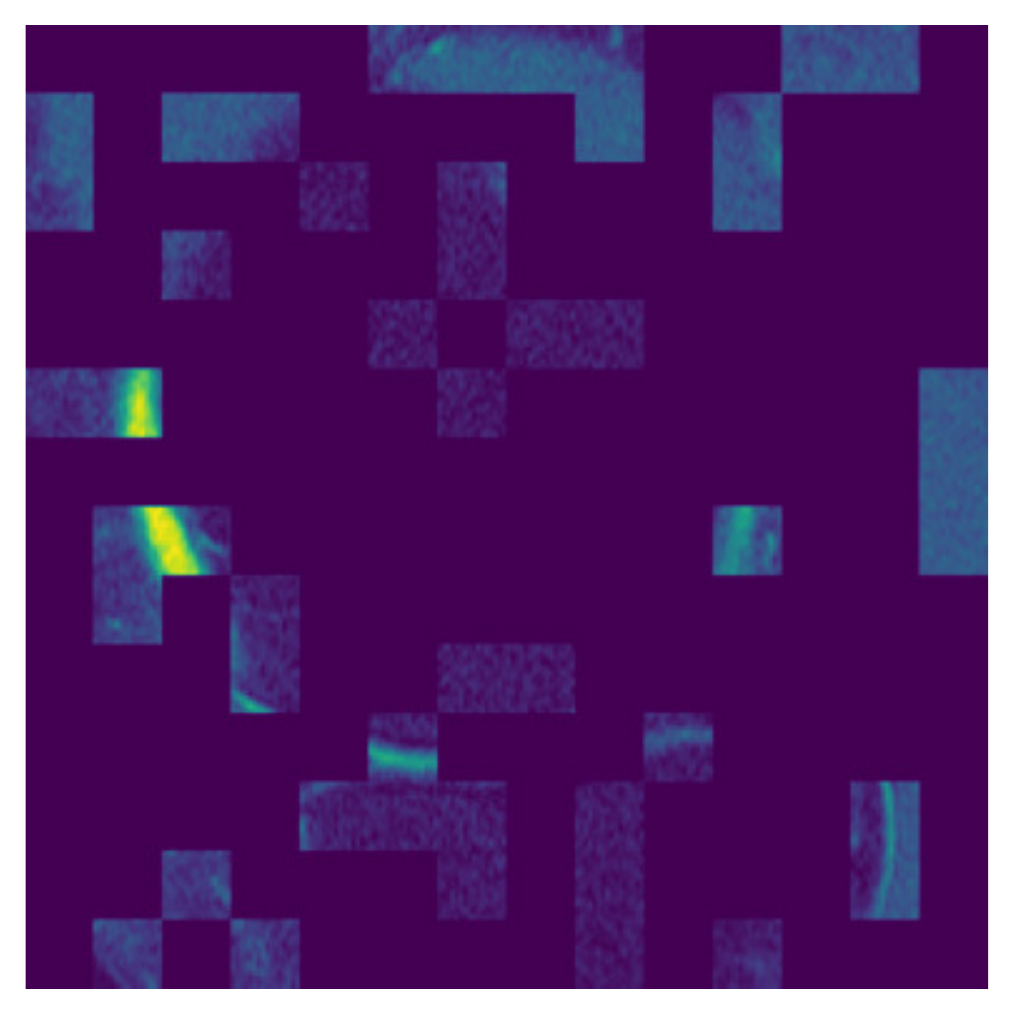}}
		\vspace{3pt}
		\centerline{\includegraphics[width=\textwidth]{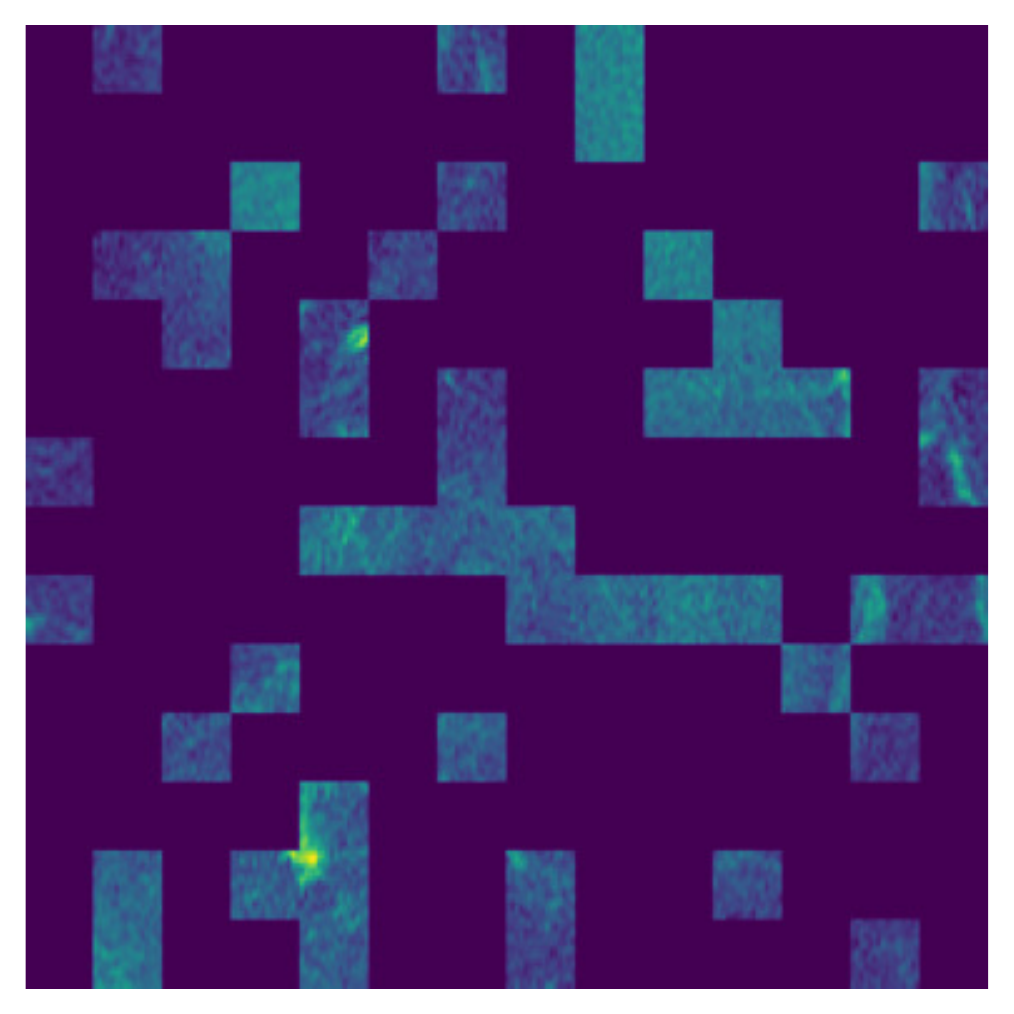}}
		\vspace{3pt}
		\centerline{\includegraphics[width=\textwidth]{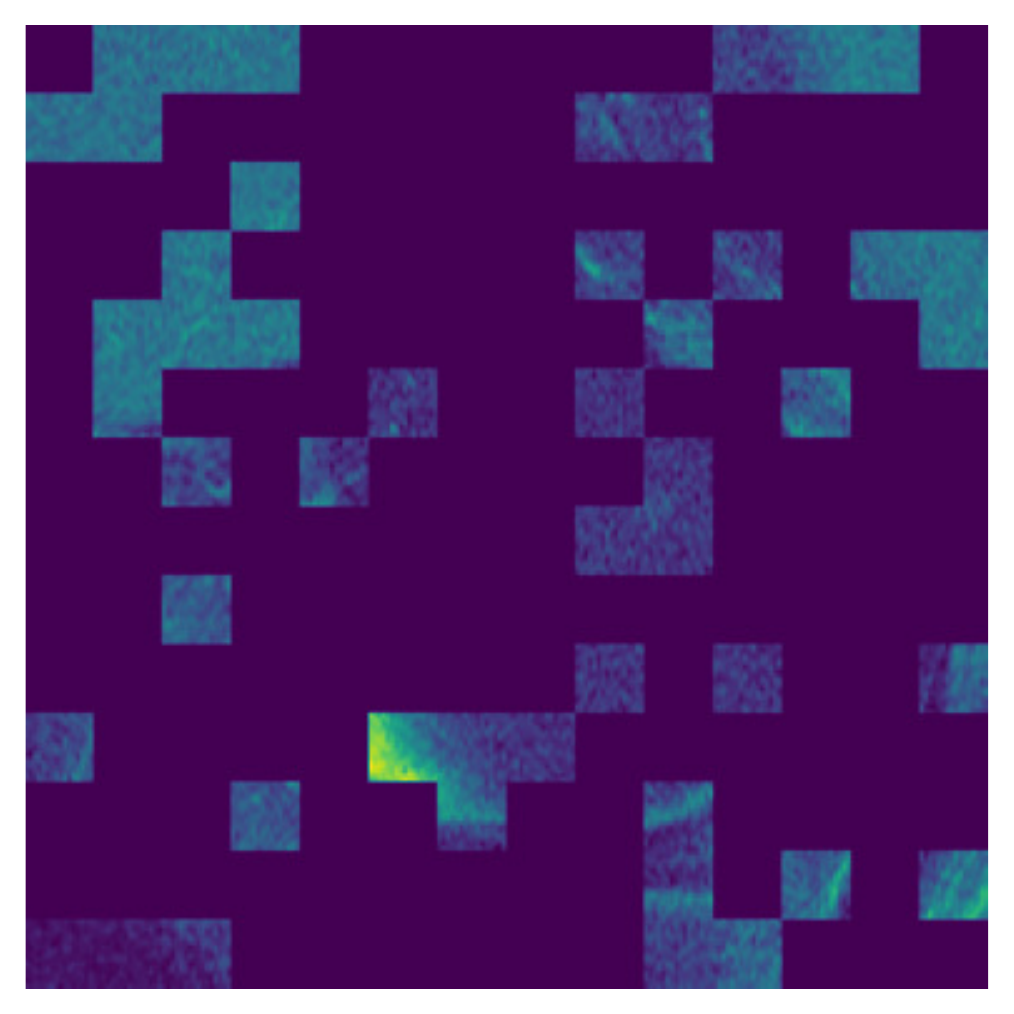}}
		\centerline{(b)}
	\end{minipage}
	\begin{minipage}{0.30\linewidth}
		\vspace{3pt}
		\centerline{\includegraphics[width=\textwidth]{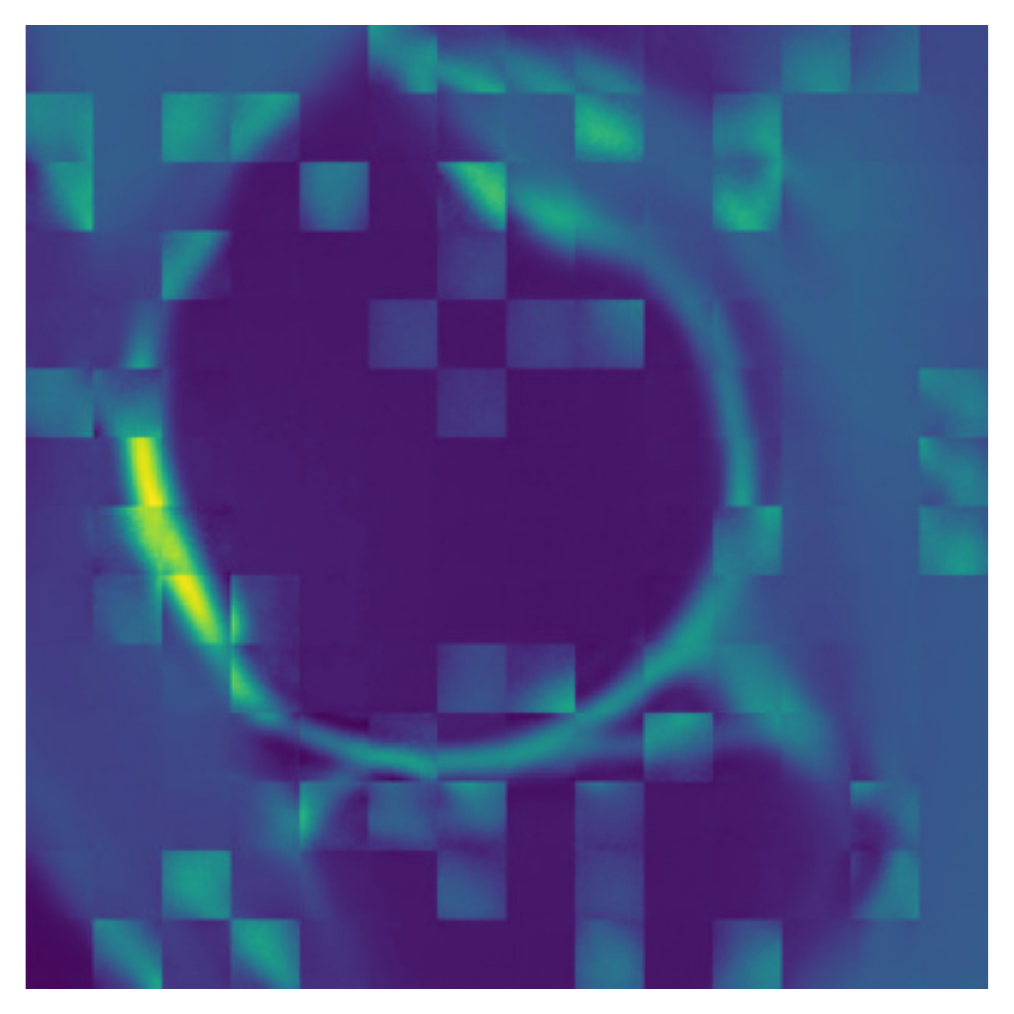}}
		\vspace{3pt}
		\centerline{\includegraphics[width=\textwidth]{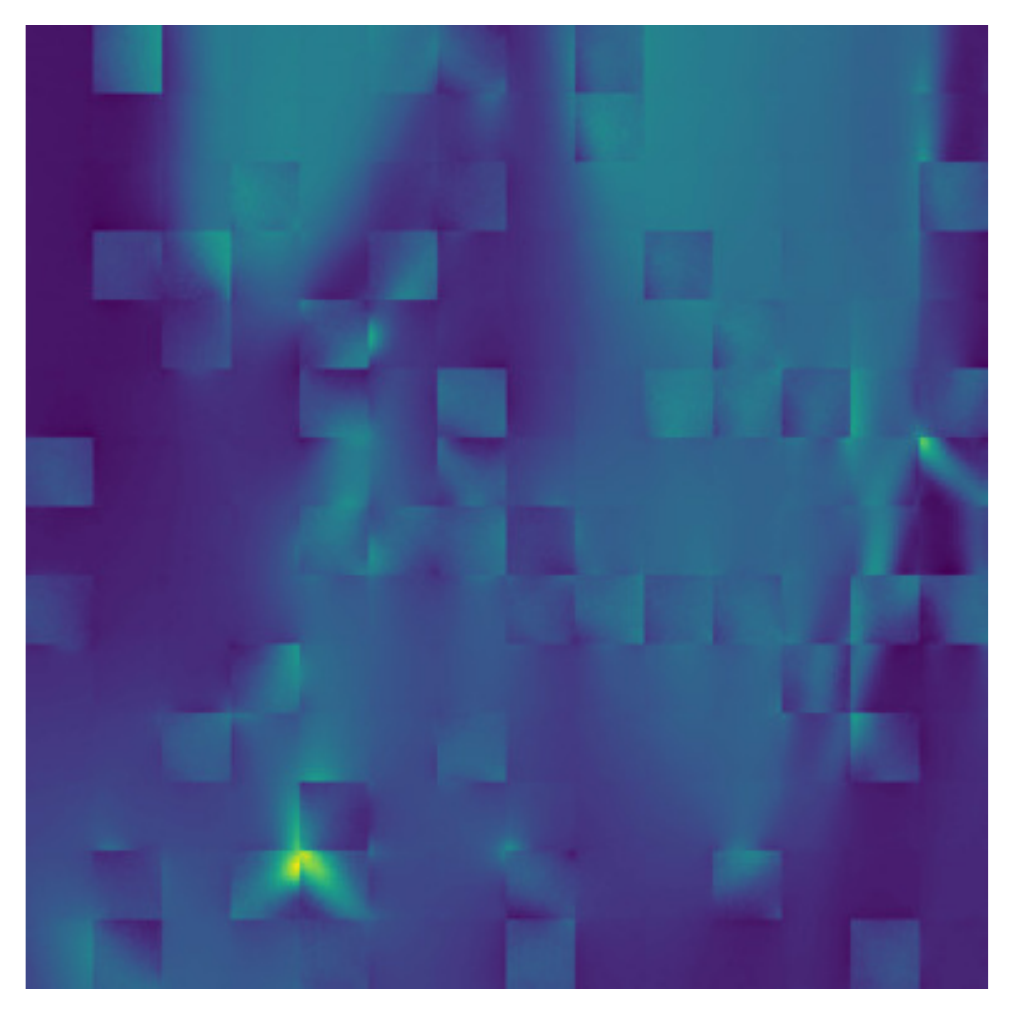}}
		\vspace{3pt}
		\centerline{\includegraphics[width=\textwidth]{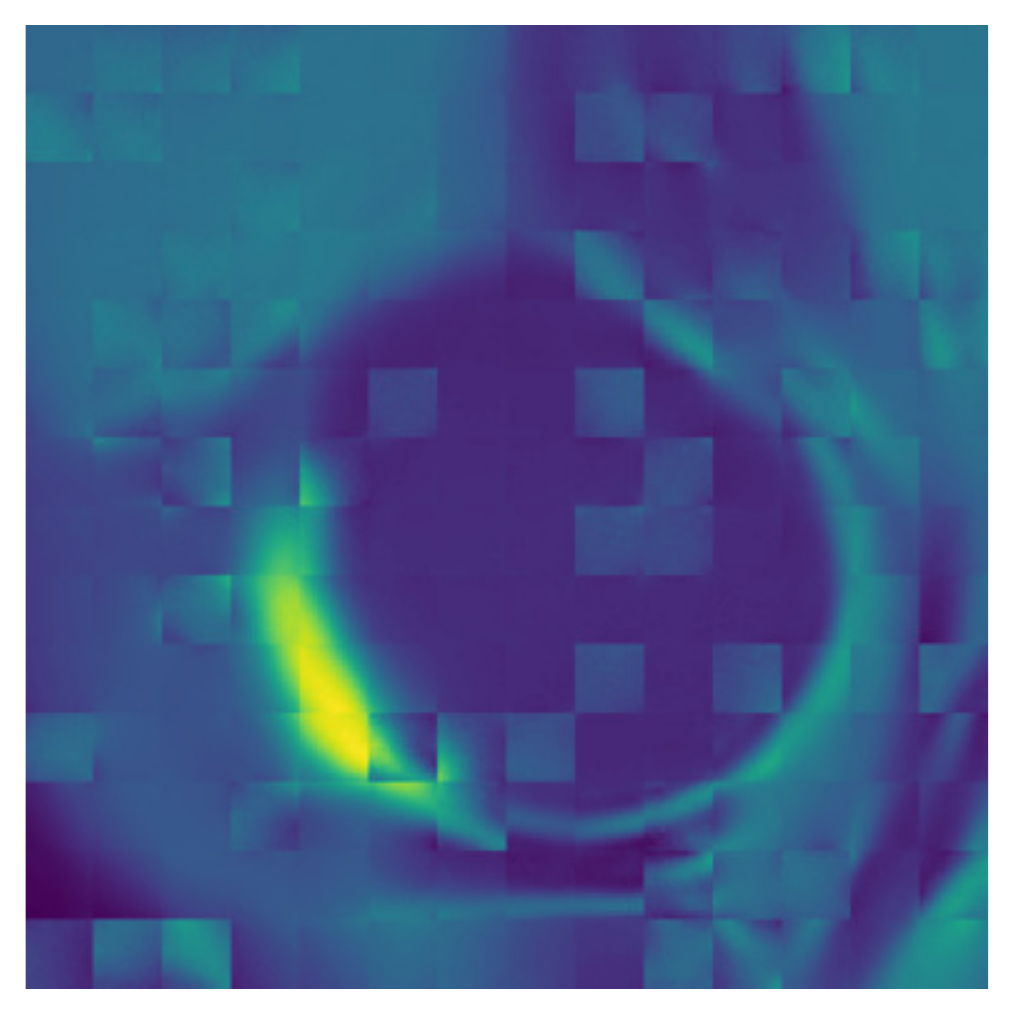}}
		\centerline{(c)}
	\end{minipage}
	\caption{Visualization of MAE reconstruction results; (a)original image (b) masked image (c) reconstruction}
\end{figure}

As shown in Figure 8, the MAE reconstruction results of the anterior and posterior horns, the body, and the slice without meniscus are shown, where the first column is the image after random masking, the second column is the original image, and the third column is the reconstruction output. On the whole, MAE can basically reconstruct the rough outline of the original image from very few image blocks, which fully shows that the pre-training process of MAE greatly improves the model's ability to understand and infer the original data, which is important for subsequent weak supervision. It is worth noting that the reconstruction results of the original visible image blocks are blurred, because the visible image blocks are not included in the calculation of the loss function, and according to the original paper of MAE, the loss function is not included in the visible image blocks so that the model has better transfer ability, so we also follow this experimental principle in our use.
\subsection{Analysis of pseudo-label generation results}
Part of the pseudo-label generation results are visualized as follows: it can be seen that with our pseudo-label generation algorithm, the generated pseudo-labels have a high similarity with the original labels, which is very important for subsequent weak supervision, because the pseudo-labels quality greatly affects the performance of the segmentation.

\begin{figure}[h]
	\centering
	\begin{minipage}{0.22\linewidth}
		\vspace{3pt}
		\centerline{\includegraphics[width=\textwidth]{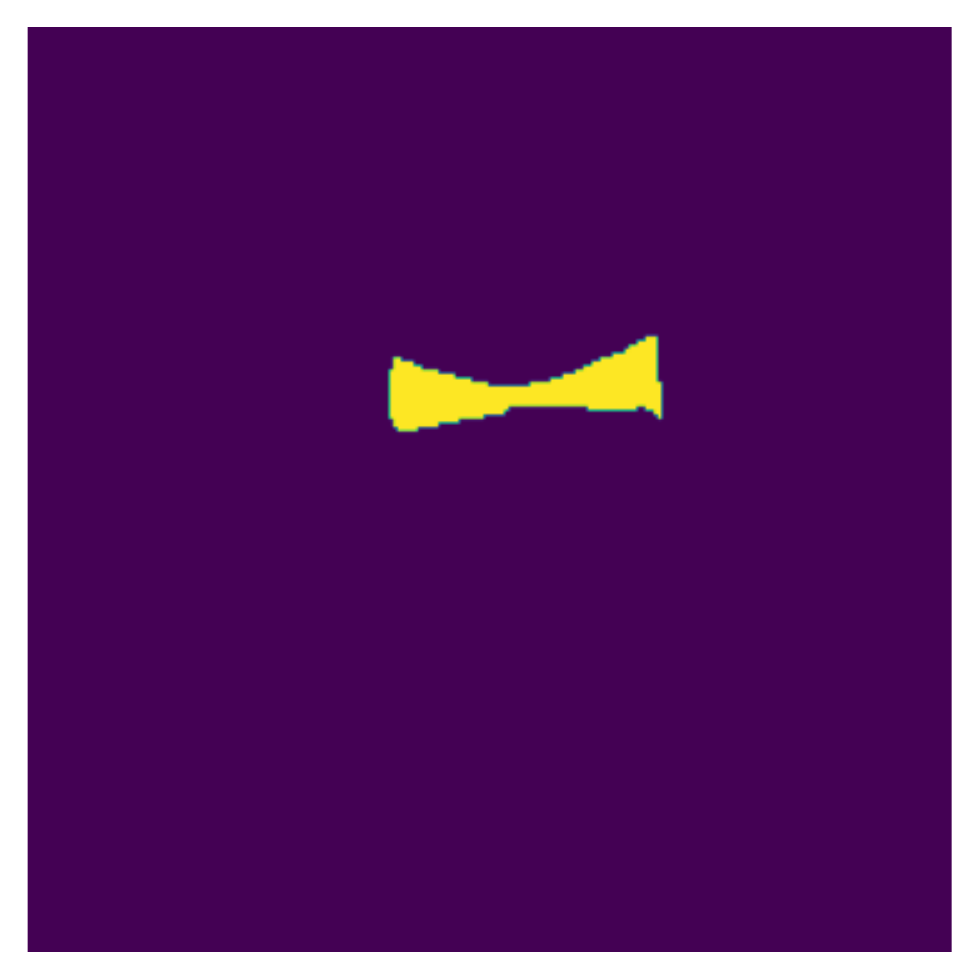}}
		\vspace{3pt}
		\centerline{\includegraphics[width=\textwidth]{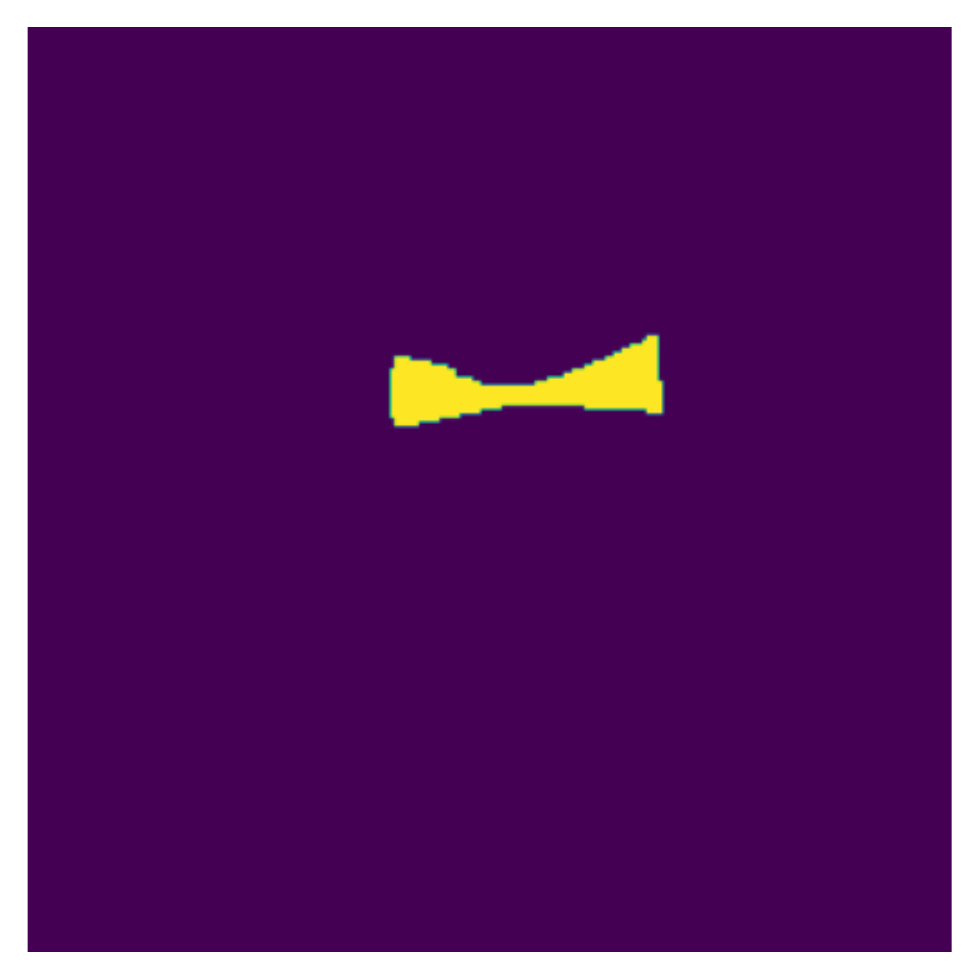}}
		
	\end{minipage}
	\begin{minipage}{0.22\linewidth}
		\vspace{3pt}
		\centerline{\includegraphics[width=\textwidth]{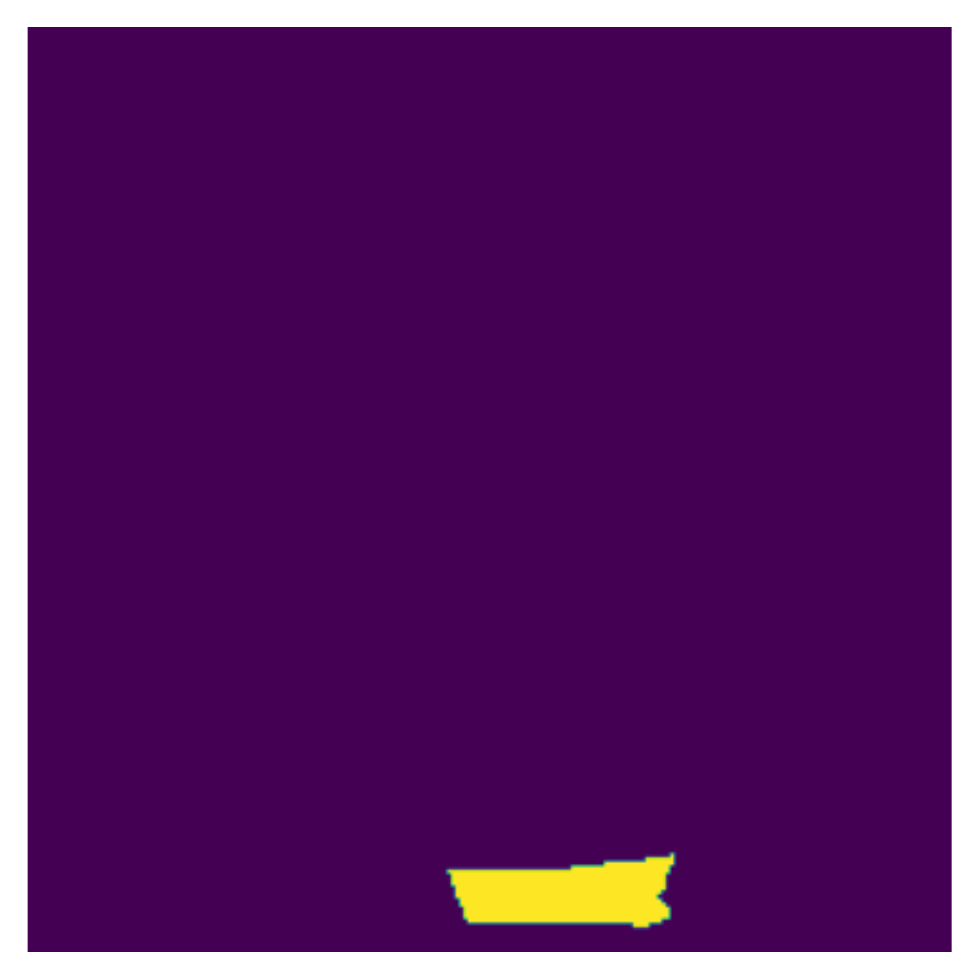}}
		\vspace{3pt}
		\centerline{\includegraphics[width=\textwidth]{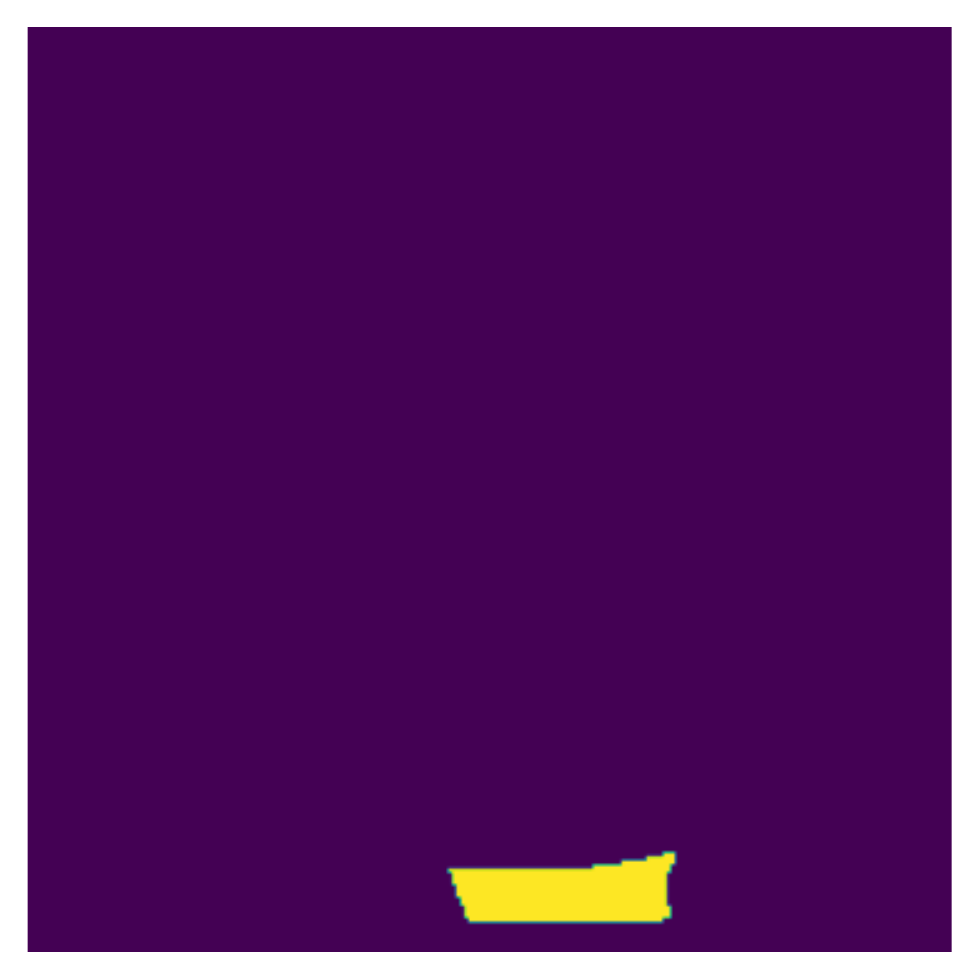}}
			
	\end{minipage}
	\begin{minipage}{0.22\linewidth}
		\vspace{3pt}
		\centerline{\includegraphics[width=\textwidth]{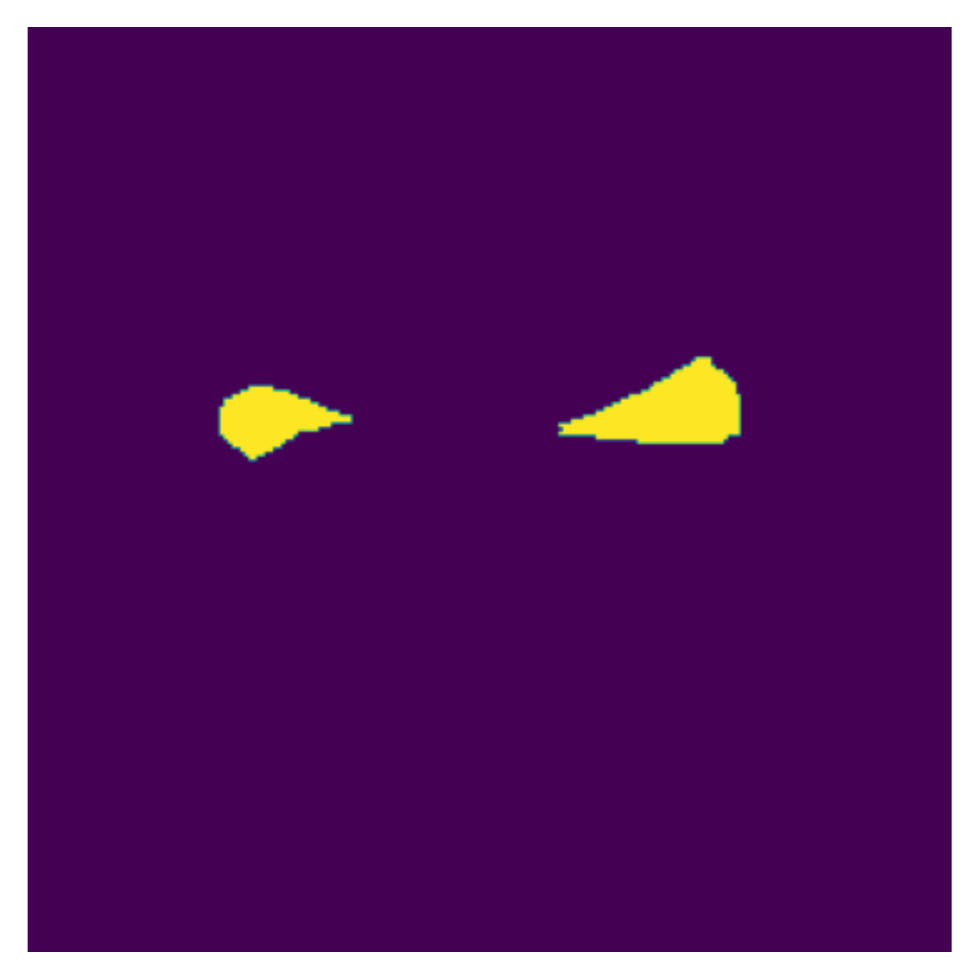}}
		\vspace{3pt}
		\centerline{\includegraphics[width=\textwidth]{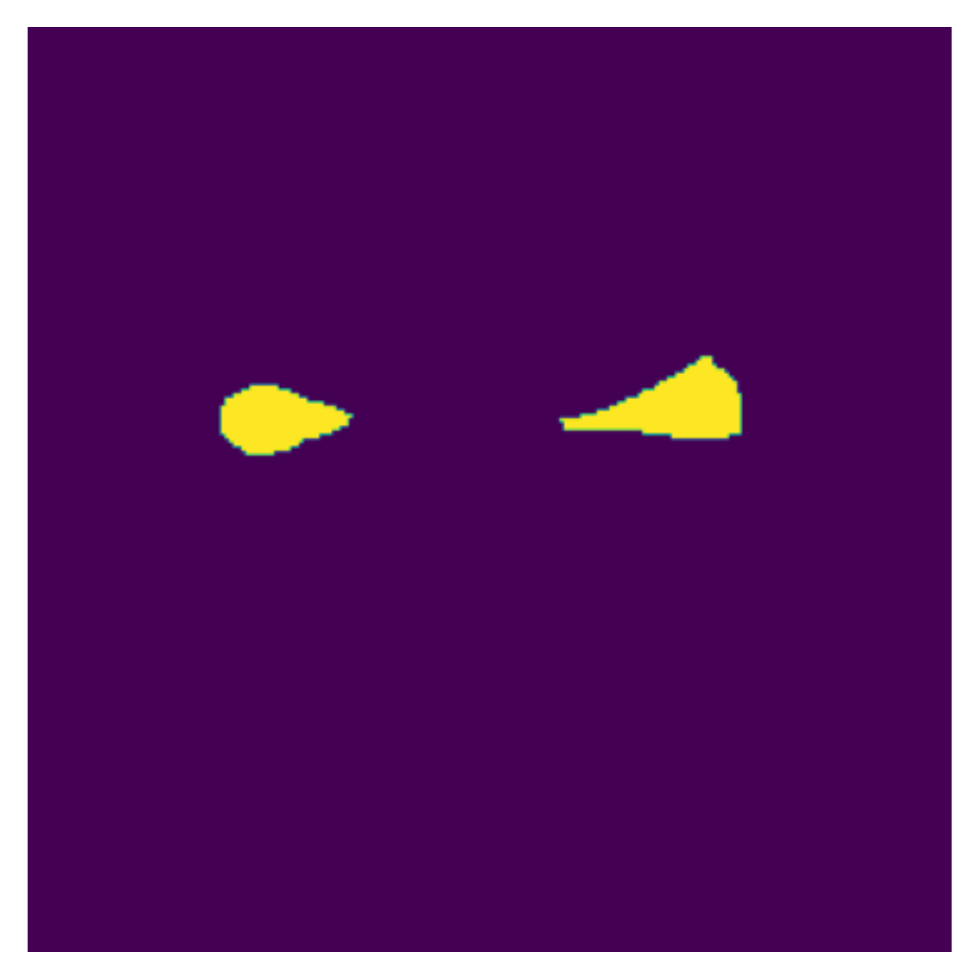}}
	
	\end{minipage}
	\begin{minipage}{0.22\linewidth}
		\vspace{3pt}
		\centerline{\includegraphics[width=\textwidth]{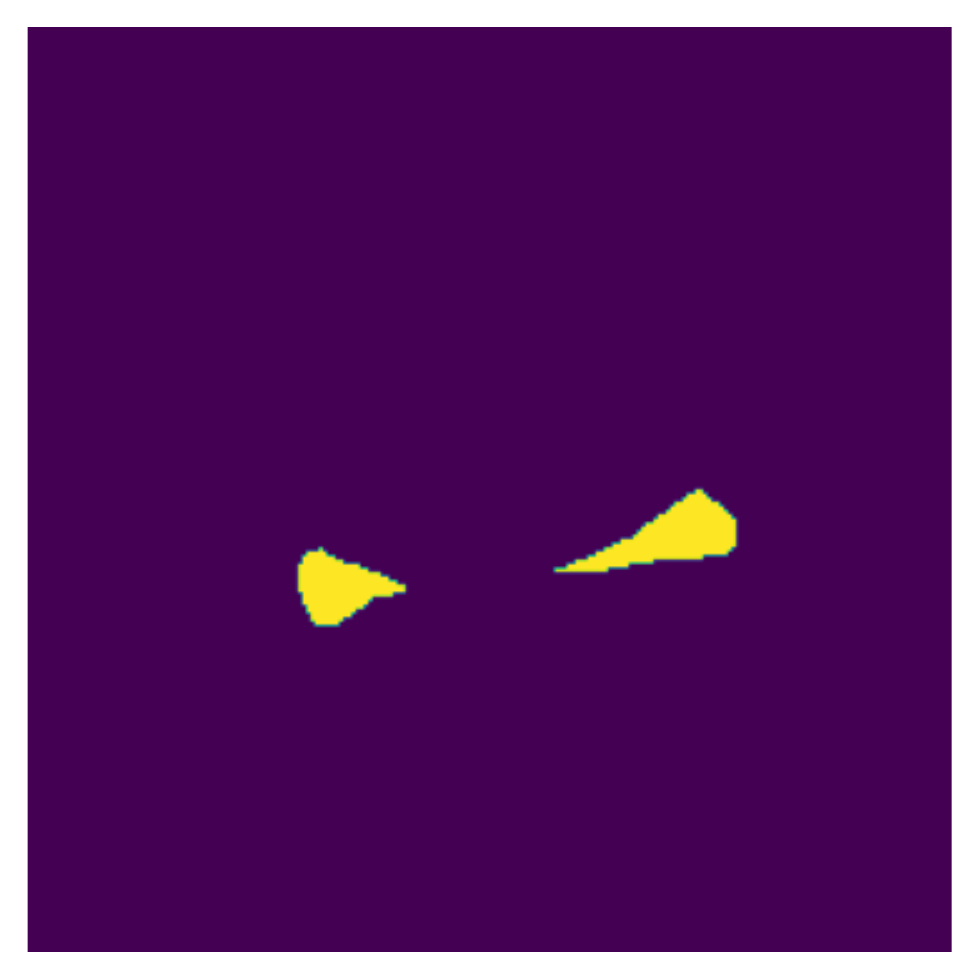}}
		\vspace{3pt}
		\centerline{\includegraphics[width=\textwidth]{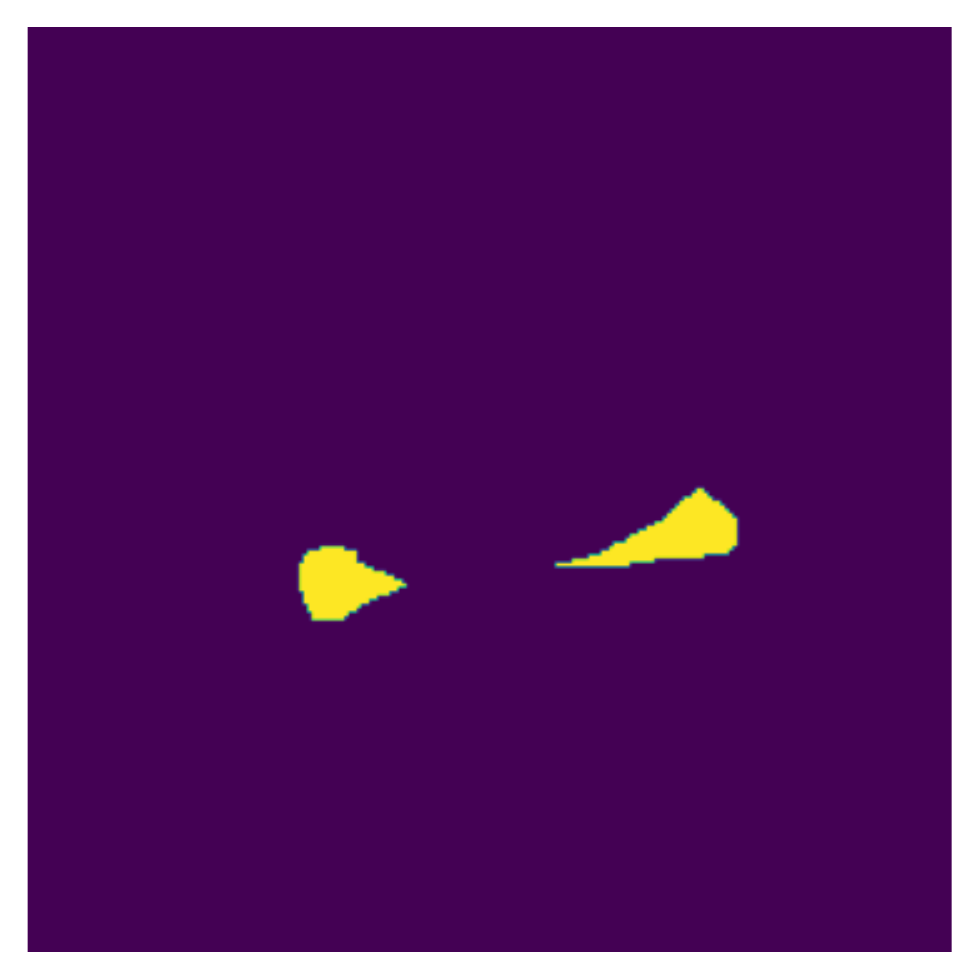}}
	
	\end{minipage}
	\caption{Comparison of pseudo-labels and real labels: the first row is the real label, and the second row is the pseudo-label}
\end{figure}

For our pseudo-label generation algorithm, we conducted a series of ablation experiments to explore the role of each part. The index used is the Dice coefficient of the pseudo-label and the real label, as shown in the following table:

\begin{table}[]
\caption{ablation experiment for pseudo-label generation algorithm}
\label{table:headings}
\centering
	\begin{tabular}{ll}
		\hline
		Method                                                   & Dice \\ \hline
		Center point growth                                      & 0.64 \\
		Backbone growth                                          & 0.80 \\
		Backbone growth + Difficult area filling                 & 0.90 \\
		Backbone growth + Difficult area filling + Edge limiting & \textbf{0.95} \\ \hline
	\end{tabular}
\end{table}

It can be seen that under the design of our pseudo-label generation algorithm, the dice coefficient of the pseudo-label and the real label is as high as 0.95, which can achieve the effect of being highly consistent with the original real label. This ablation experiment also proves that our multiple steps are effective measures, and the best results can be achieved in combination.

\subsection{Performance Analysis of Self-Supervised and Weakly Supervised Segmentation}

From Table 3, it can be seen that using transunet as the backbone and using MAE for self-supervision can achieve the best results; the segmentation results of VIT+UNETR also verify the effectiveness of MAE self-supervision from the side. In transunet and VIT+UNETR, MAE can significantly improve downstream tasks whether it is weakly supervised or fully supervised. In addition, the comparison of the two transfer methods also shows that the transfer of MAE is not only limited to the encoder with the same architecture, but can also be applied to other position. We also observe that the model converges faster than training from scratch during MAE-based transfer training, which also verifies the effectiveness of self-supervision on the meniscus from the side.

\begin{table}[]
	\caption{Segmentation Dice coefficient (\%) results of fully supervised and weakly supervised datasets}
	\label{table:headings}
	\centering
	\begin{tabular}{ccccc}
		\hline
		\multirow{2}{*}{Method} & \multicolumn{2}{c}{Fully Supervised} & \multicolumn{2}{c}{Weakly Supervised} \\ \cline{2-5} 
		& MAE pretrain          & None         & MAE pretrain           & None         \\ \hline
		Swinunet                & --                    & 80.23        & --                     & 78.85        \\
		VIT+UNETR(2d)           & 83.62                 & 82.44        & 81.92                  & 80.75        \\
		Res50+unet              & --                    & 85.88        & --                     & 84.39        \\
		Transunet               & \textbf{87.25}        & 86.77        & \textbf{85.42}         & 84.98        \\ \hline
	\end{tabular}
\end{table}

For the choice of backbone, the comparison between res50+unet and transunet, as well as the comparison between swinunet and transunet can prove that in the meniscus dataset, the pure transformer or pure CNN structure is not suitable, but the transformer and the the fusion of CNN architecture can combine the local attention mechanism of convolution and the global attention mechanism of transformer to achieve better performance.

By comparing the results of weak supervision and full supervision, we can see the advantages brought by our pseudo-label generation algorithm. In many experiments, the gap between full supervision and weak supervision remained at about 1 to 2 percentage points; With the help of self-supervision, the combination of self-supervision and weak supervision can almost approach the same performance as purely using full supervision(86.77 vs 85.42 in transunet; 82.44 vs 81.92 in VIT+UNETR).

In addition, we also conduct experiments on some SOTA algorithm frameworks based on rectangular boxes, trying to redesign the loss function and improve the network architecture based on rectangular boxes, and compare them with the pure segmentation network. All the compared methods use MAE for pretraining, transunet as the backbone network and conduct weakly supervised training and testing; the results are shown in the following table 4. It can be seen that in the knee joint data set, the natural image method may not be suitable, and the complex architecture design may further increase the difficulty of optimization.

\begin{table}[]
\caption{Comparison of weakly supervised architecture based on rectangular box and pure segmentation architecture}
\label{table:headings}
\centering
	\begin{tabular}{ll}
		\hline
		Method      & Dice   \\ \hline
		Box2Seg     & 0.8251 \\
		BCM         & 0.8308 \\
		Pure Segnet & \textbf{0.8542} \\ \hline
	\end{tabular}
\end{table}

Since we initially used only 70 cases of data in the original training set for self-supervised training, considering that there is still a large amount of unlabeled data in the database that can be utilized, we used additional 1300 cases of data to perform MAE self-supervised training again to examine the impact of dataset changes on downstream segmentation tasks. The results are as follows: It can be seen that the expansion of the dataset used by MAE self-supervision enables the model to have better representational capabilities than only on the training set, which can bring a small performance improvement based on the result of MAE pretraining on 70 cases.

\begin{table}[]
	\caption{The effect(use dice coefficient (\%) as metric) of the size of MAE pre-training dataset on the segmentation task}
	\label{table:headings}
	\centering
	\begin{tabular}{ccccc}
		\hline
		\multirow{2}{*}{Method} & \multicolumn{2}{c}{Fully Supervised} & \multicolumn{2}{c}{Weakly Supervised} \\ \cline{2-5} 
		& MAE(1300)    & MAE(70)   & MAE(1300)    & MAE(70)    \\ \cline{2-5} 
		VIT+UNETR(2d)           & 83.91              & 83.62           & 82.29              & 81.92            \\
		Transunet               & \textbf{87.40}     & 87.25           & \textbf{85.59}     & 85.42            \\ \hline
	\end{tabular}
\end{table}

\subsection{Visualization of segmentation results}

According to the visualization of the segmentation results in figure 10, it can be seen that the segmentation results obtained based on the self-supervised and weakly supervised algorithms have good segmentation positioning and edge perception capabilities, and show excellent performance in some difficult segmentation cases, which are better than other methods.

\begin{figure*}[h]
	\centering
	\begin{minipage}{0.18\linewidth}
		\vspace{3pt}
		\centerline{\includegraphics[width=\textwidth]{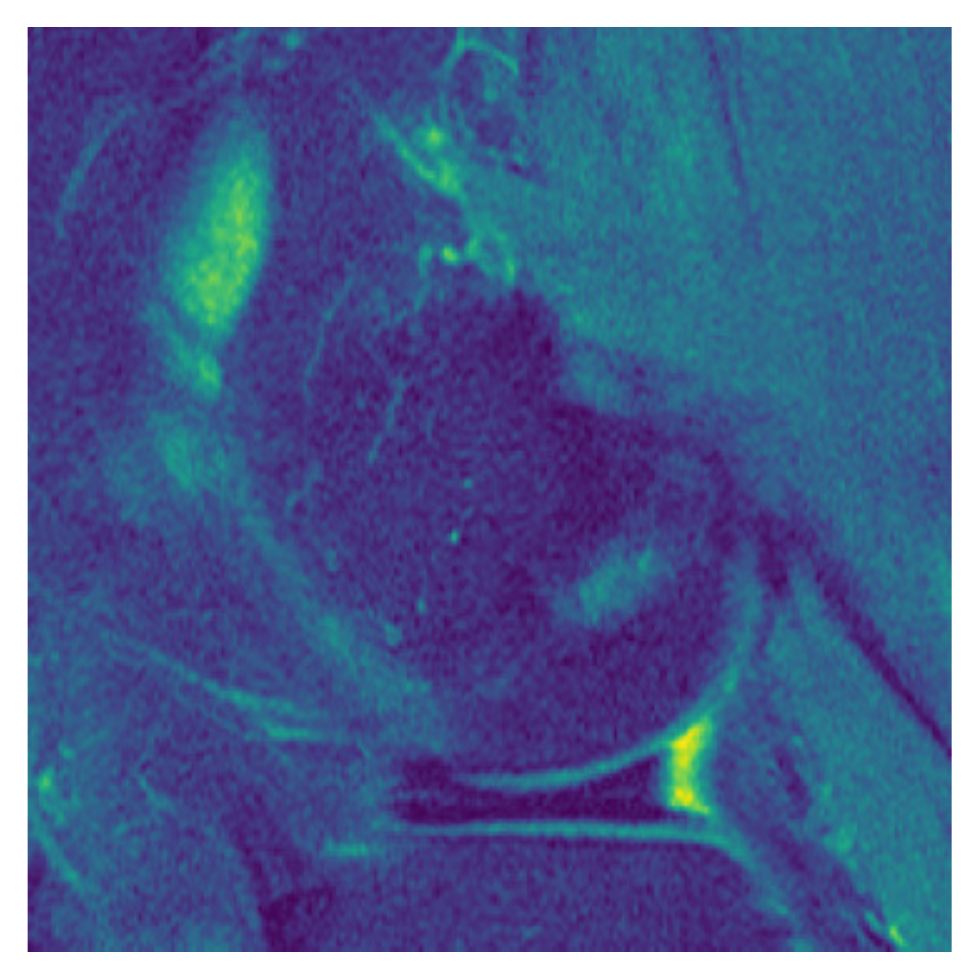}}
		\vspace{3pt}
		\centerline{\includegraphics[width=\textwidth]{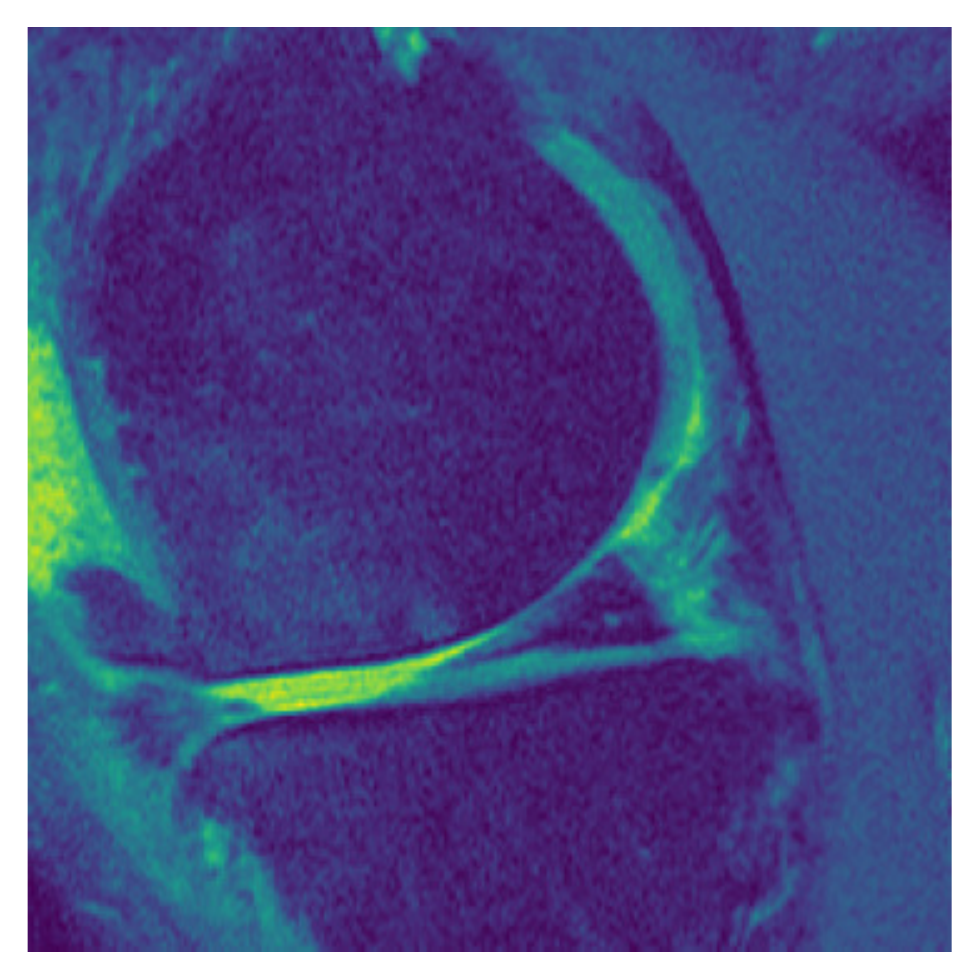}}
		\vspace{3pt}
		\centerline{\includegraphics[width=\textwidth]{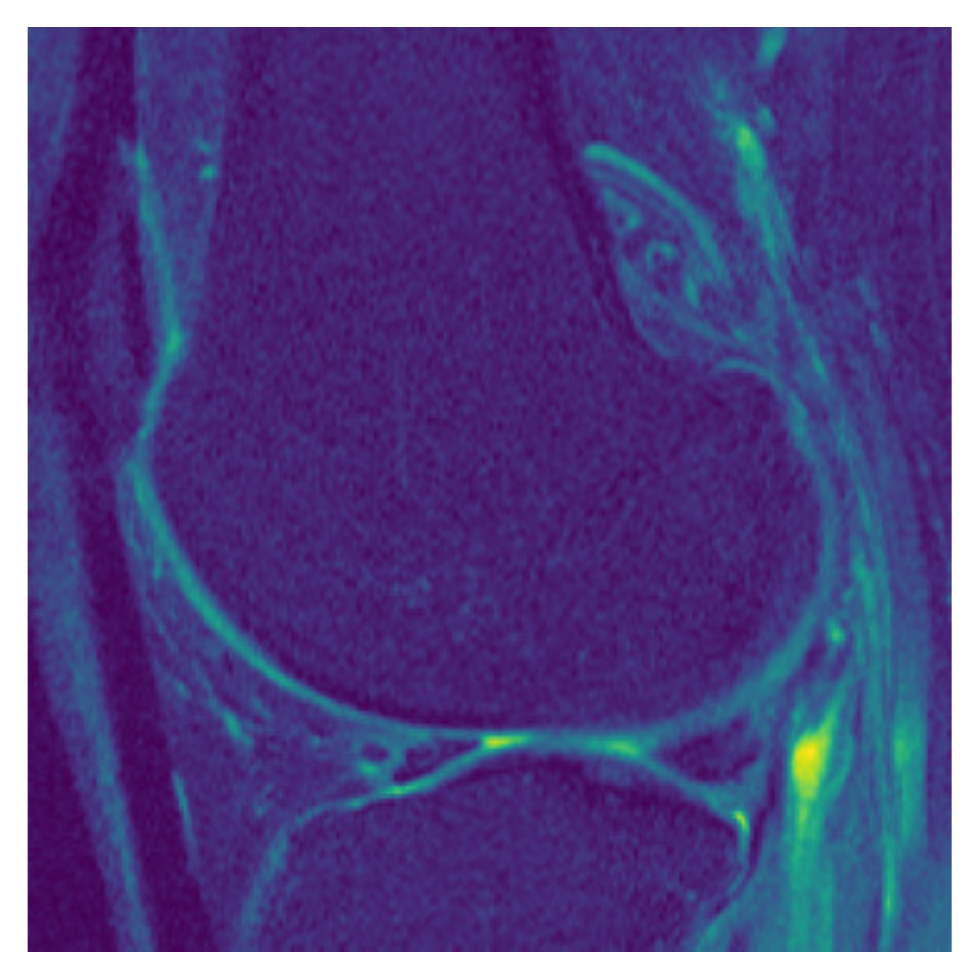}}
		\centerline{(a)}
	\end{minipage}
	\begin{minipage}{0.18\linewidth}
		\vspace{3pt}
		\centerline{\includegraphics[width=\textwidth]{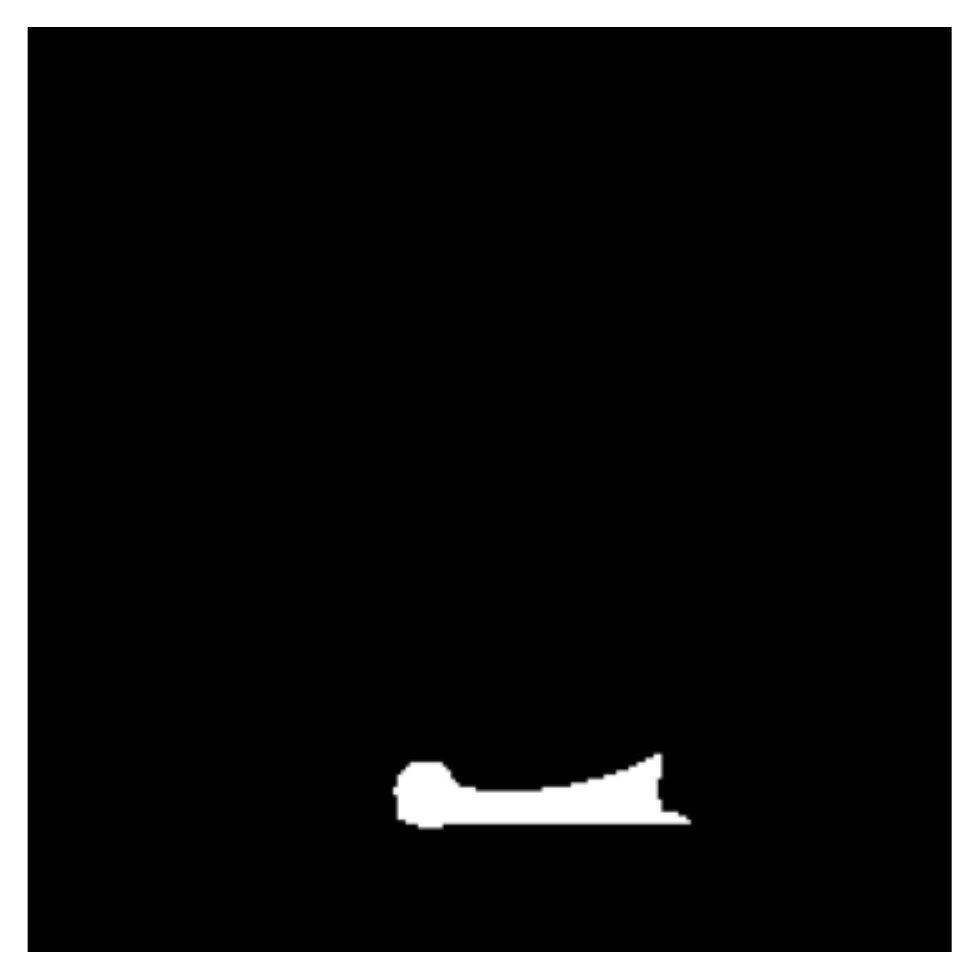}}
		\vspace{3pt}
		\centerline{\includegraphics[width=\textwidth]{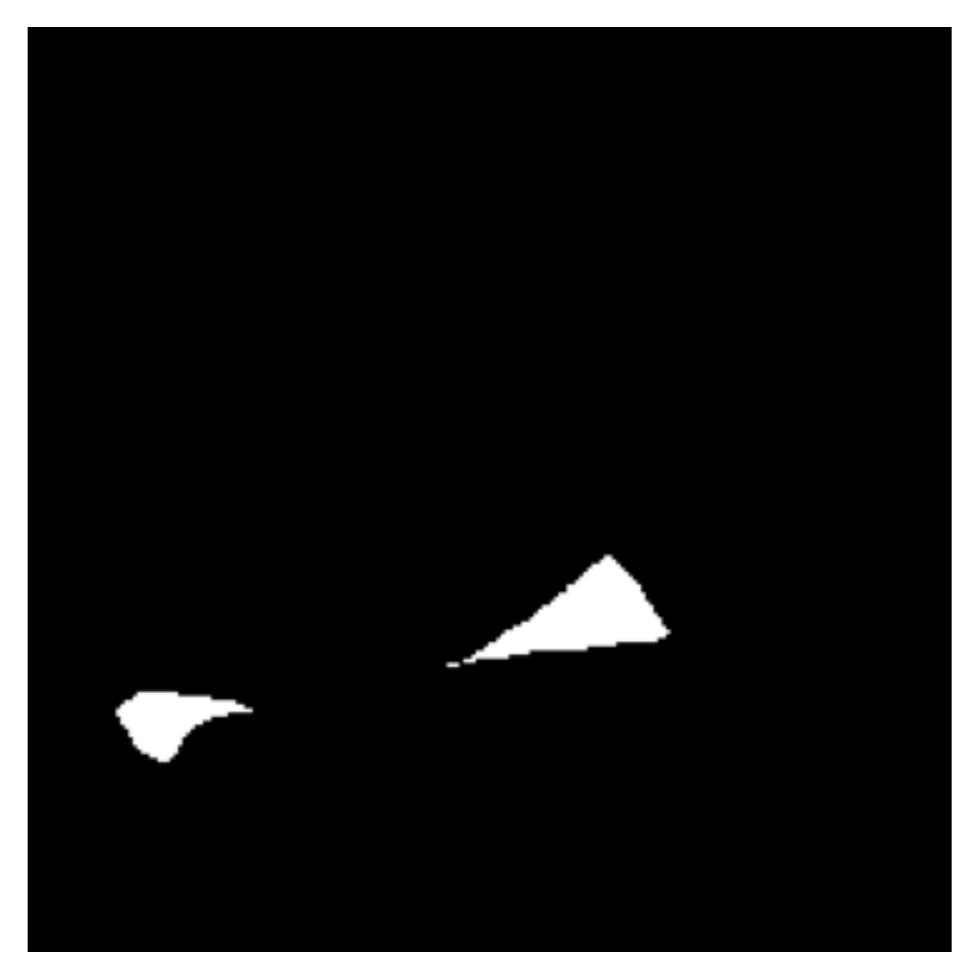}}
		\vspace{3pt}
		\centerline{\includegraphics[width=\textwidth]{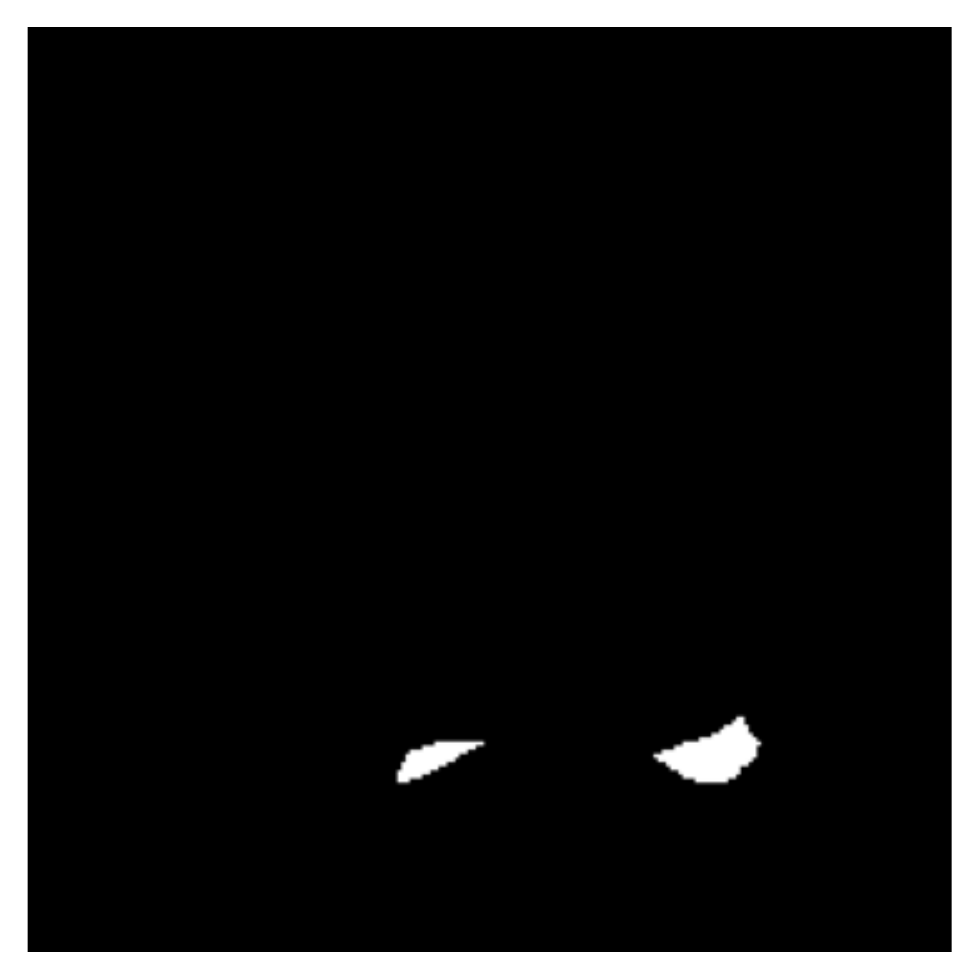}}
		\centerline{(b)}
	\end{minipage}
	\begin{minipage}{0.18\linewidth}
		\vspace{3pt}
		\centerline{\includegraphics[width=\textwidth]{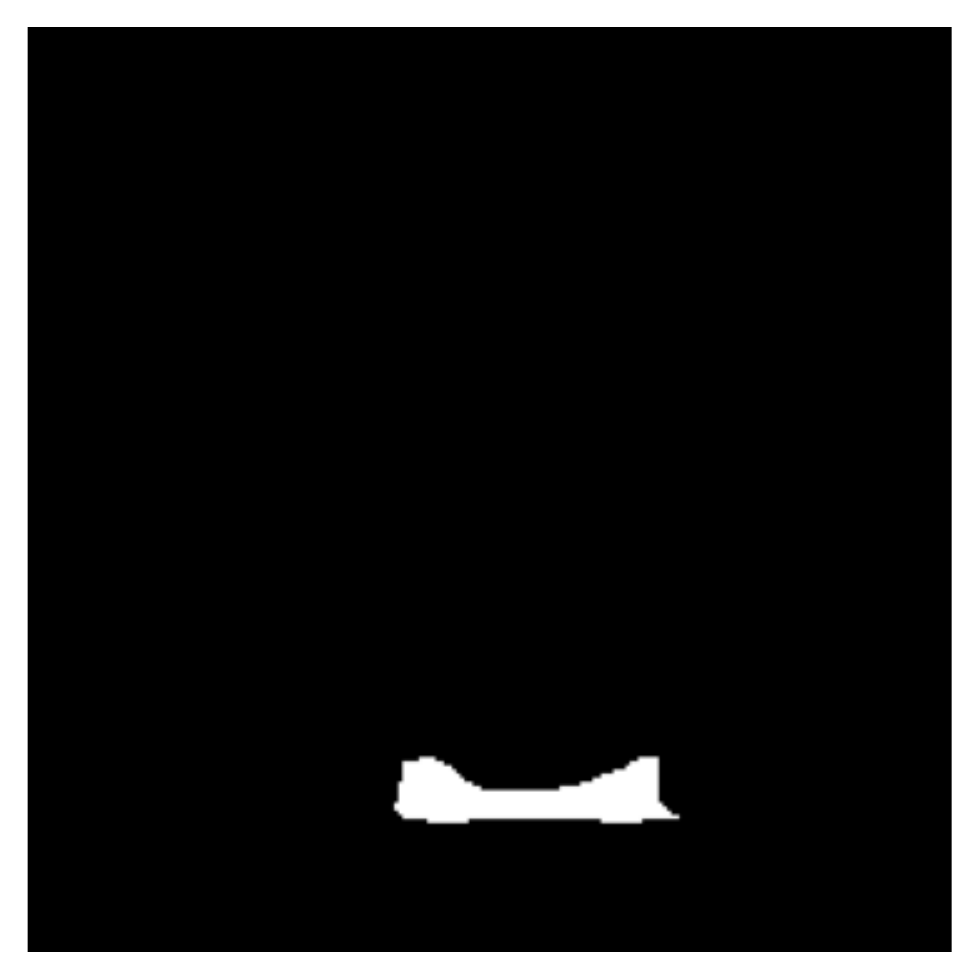}}
		\vspace{3pt}
		\centerline{\includegraphics[width=\textwidth]{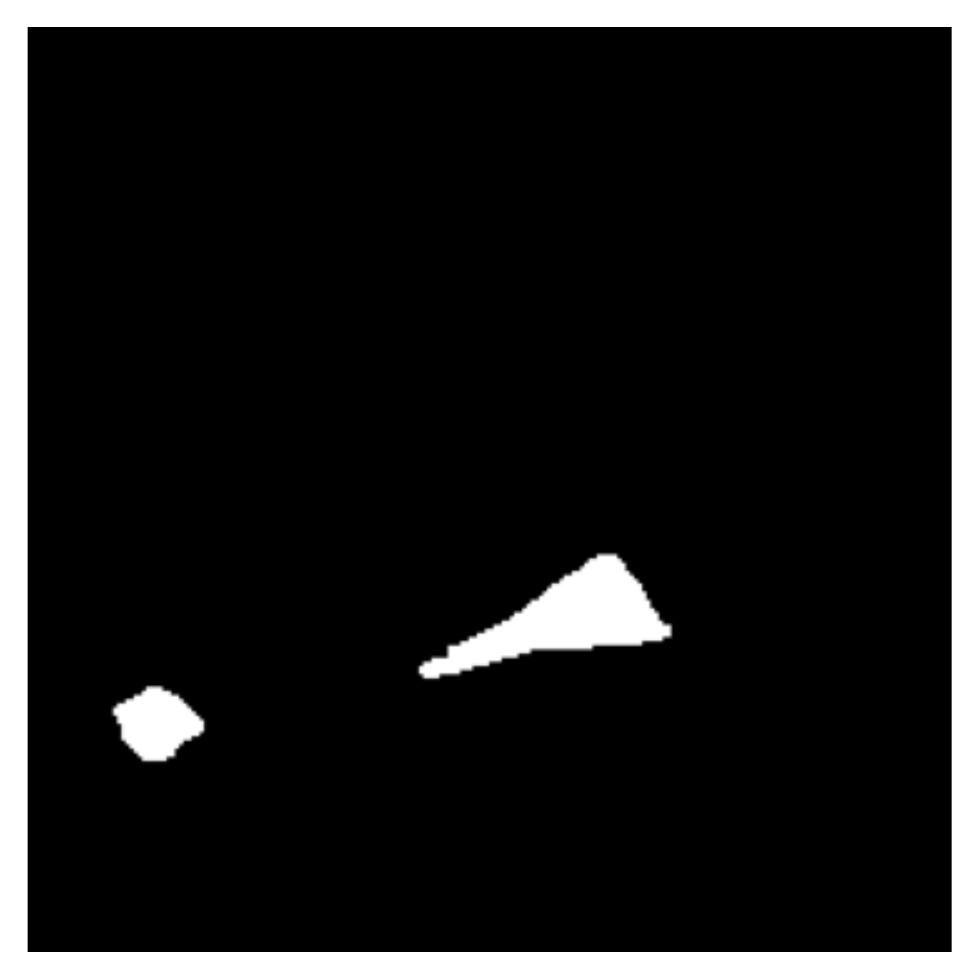}}
		\vspace{3pt}
		\centerline{\includegraphics[width=\textwidth]{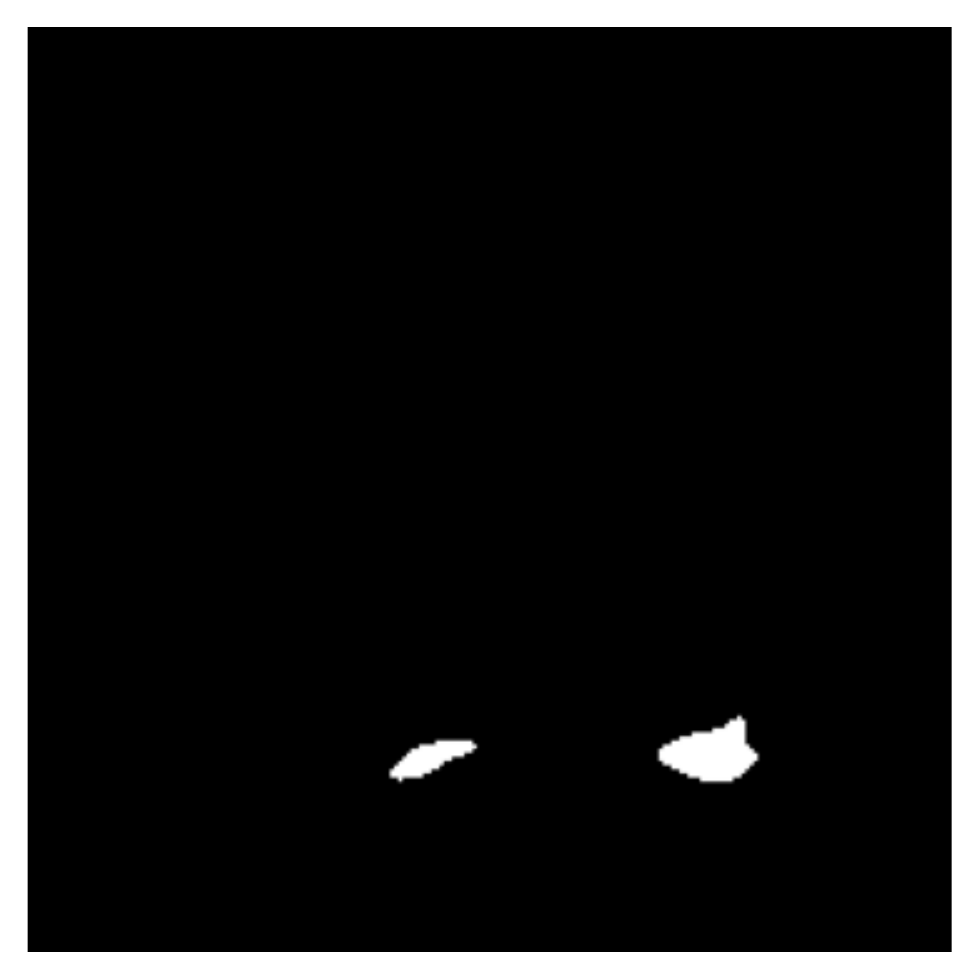}}
		\centerline{(c)}
	\end{minipage}
	\begin{minipage}{0.18\linewidth}
		\vspace{3pt}
		\centerline{\includegraphics[width=\textwidth]{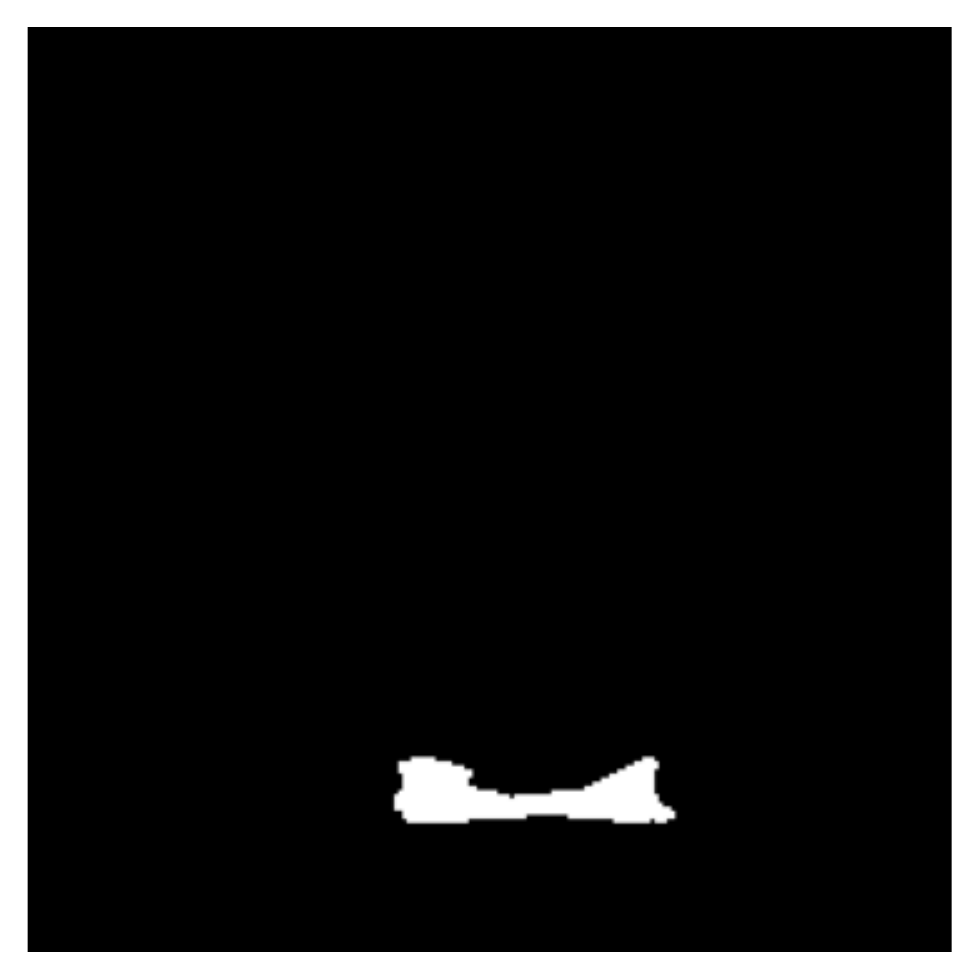}}
		\vspace{3pt}
		\centerline{\includegraphics[width=\textwidth]{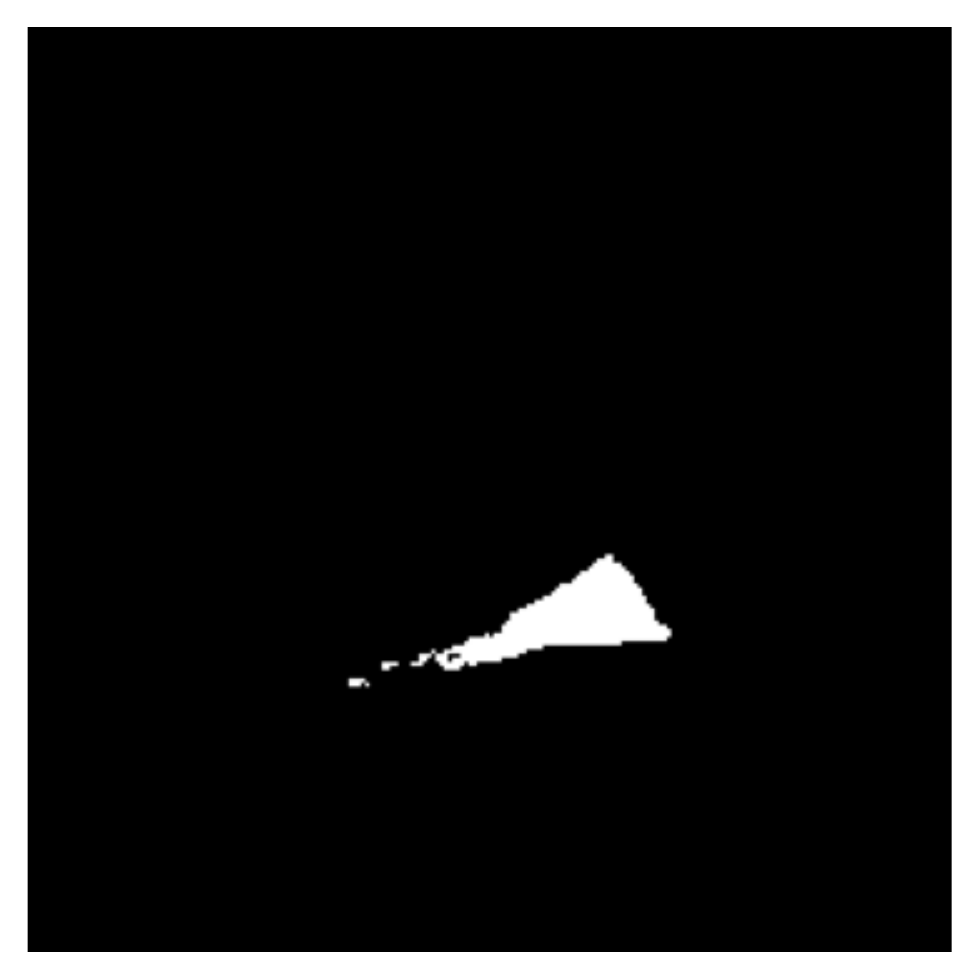}}
		\vspace{3pt}
		\centerline{\includegraphics[width=\textwidth]{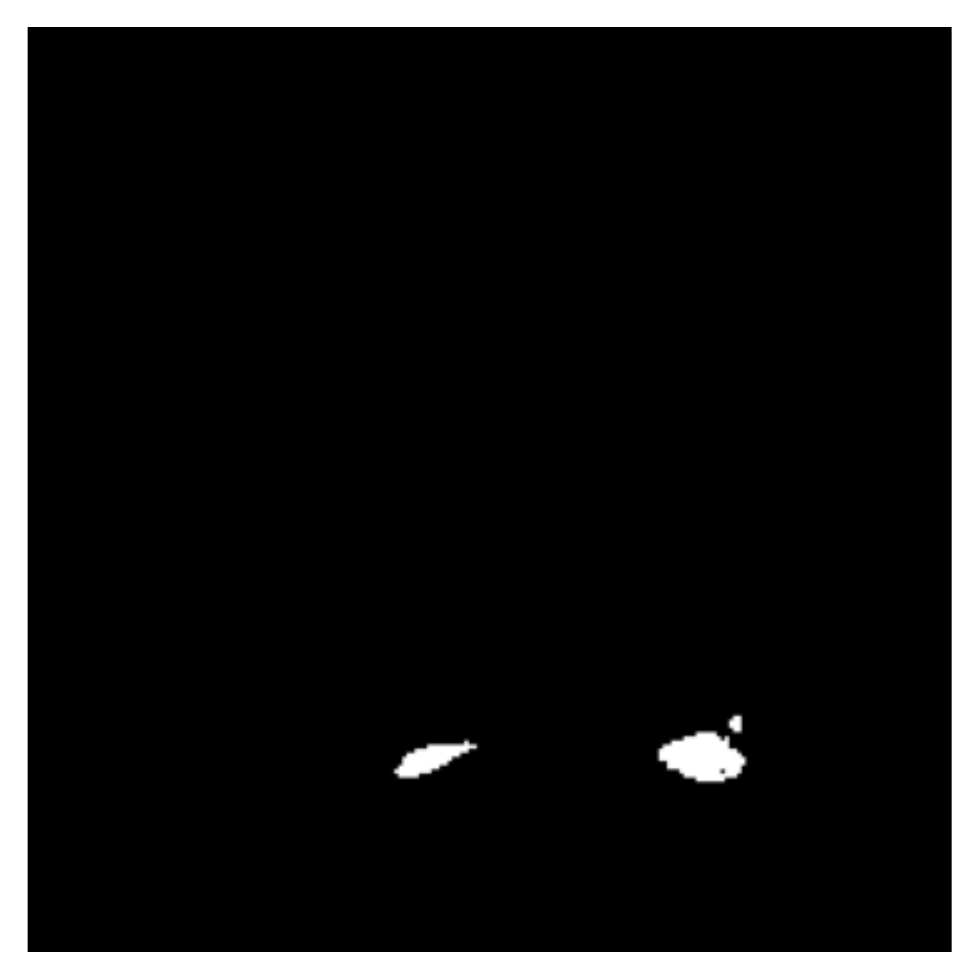}}
		\centerline{(d)}
	\end{minipage}
	\begin{minipage}{0.18\linewidth}
		\vspace{3pt}
		\centerline{\includegraphics[width=\textwidth]{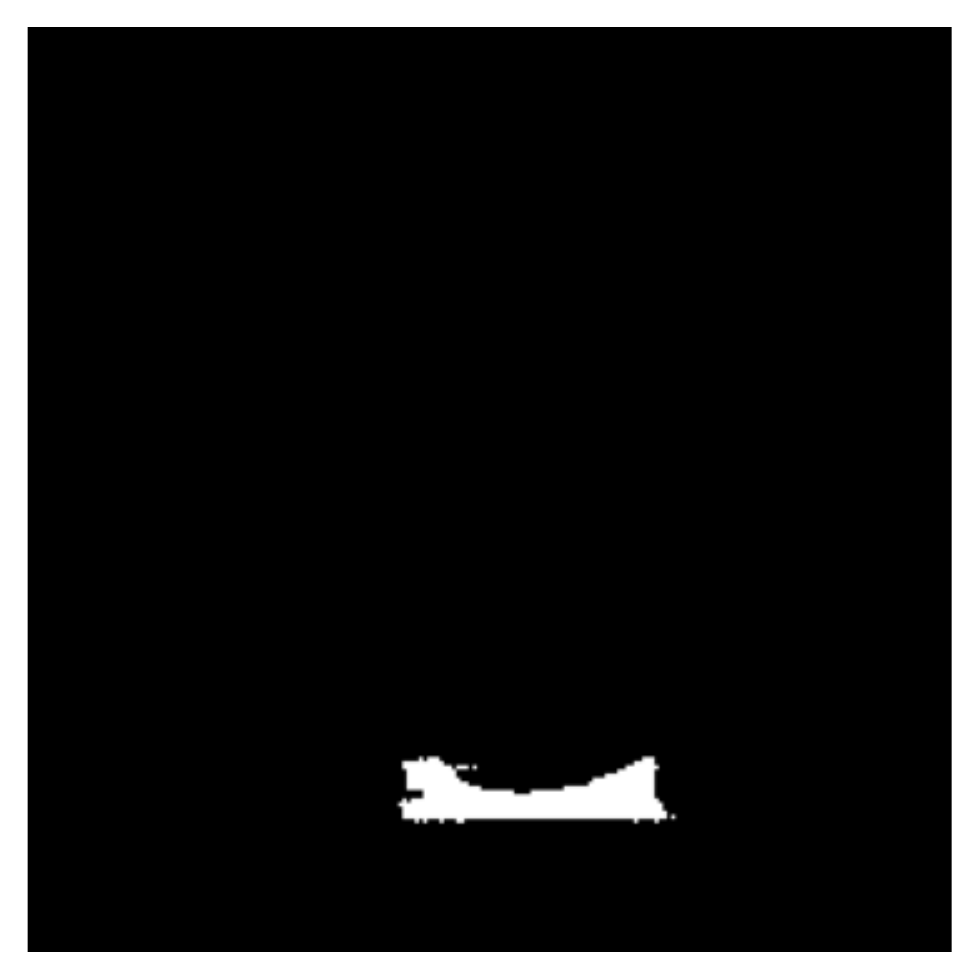}}
		\vspace{3pt}
		\centerline{\includegraphics[width=\textwidth]{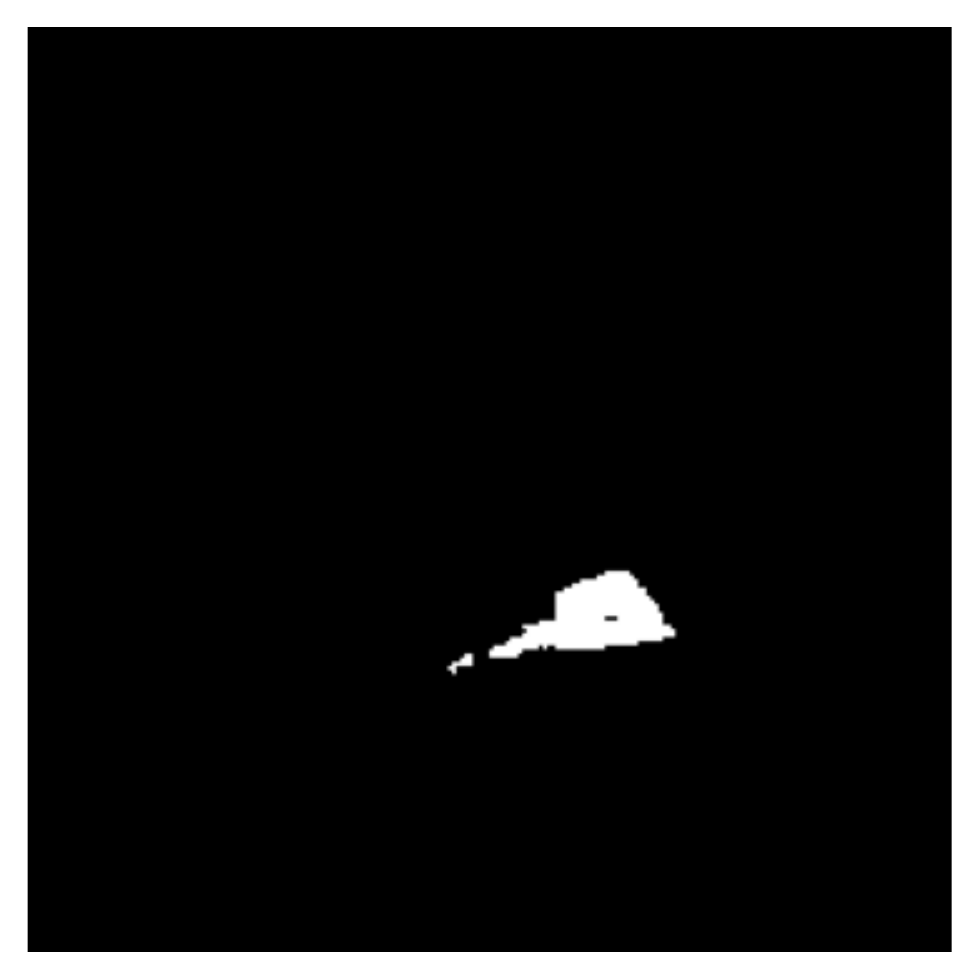}}
		\vspace{3pt}
		\centerline{\includegraphics[width=\textwidth]{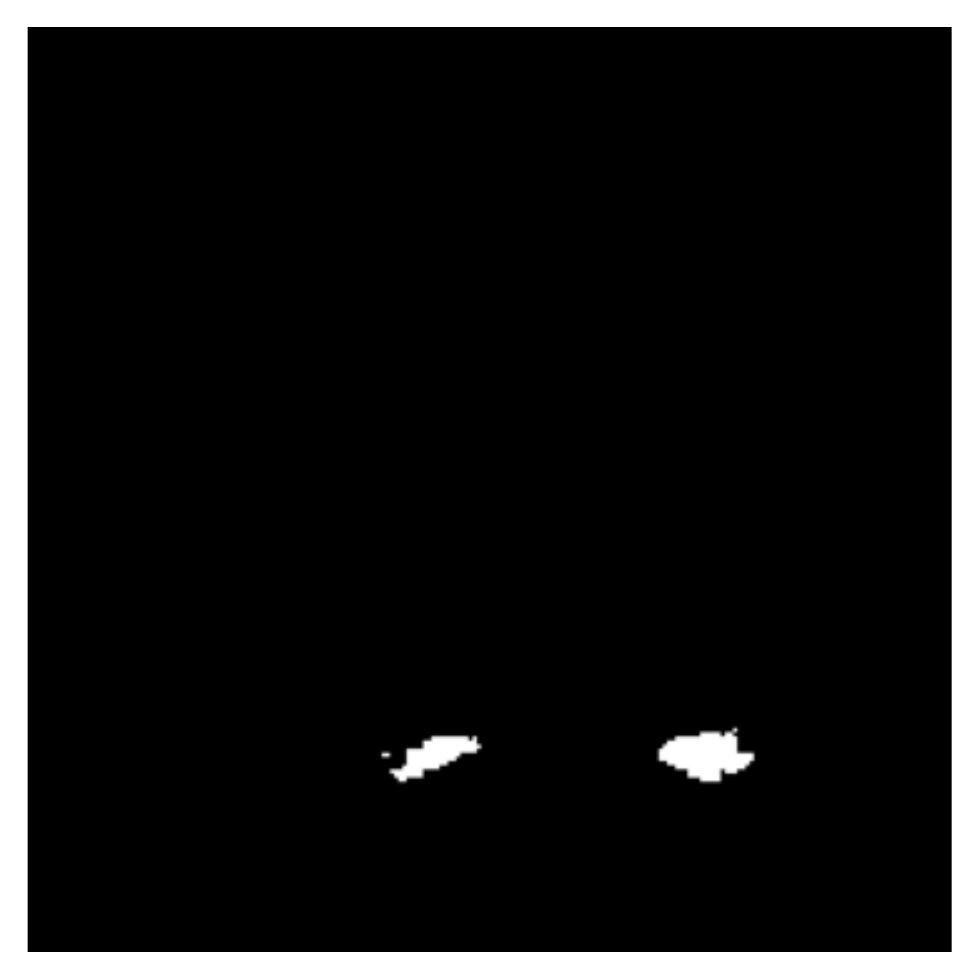}}
		\centerline{(e)}
	\end{minipage}
	\caption{Visualization of segmentation results Note: the first column (a) is the original image; the second column (b) is the real label; the third column (c) represents the result of transunet + MAE + weakly supervised; the fourth column (d) represents the result of res50unet + weakly supervised; the fifth column (d) represents the result of swinunet + weakly supervised}
\end{figure*}

\section{Conclusion}
In this paper, we apply the MAE self-supervised training to the knee joint dataset, and verify the feasibility and effectiveness of its different transfer methods in the segmentation task. Besides, we propose a weakly supervised meniscus paradigm based on the combination of points and lines, and use the backbone-based region growing algorithm to complete the generation of pseudo-labels. Finally, we conduct weakly supervised training based on the transfer weights and pseudo-labels. The sufficient experimental results verify that our algorithm can achieve almost the same performance as fully supervision. We hope that the weakly supervised paradigm based on points and lines proposed in this paper can inspire attention to weakly supervised label design; in future work, we will explore self-supervised, weakly supervised in the classification and detection tasks based on knee joint medical images and try to improve the adaptability of the algorithm in 3D medical images using the spatial information.

\bibliographystyle{splncs04}
\bibliography{egbib}

\end{document}